\documentclass{article}
\PassOptionsToPackage{numbers,sort&compress}{natbib}

\usepackage{microtype}
\usepackage{graphicx}
\usepackage{booktabs} %

\usepackage[colorlinks,citecolor=blue,urlcolor=black,linkcolor=blue,pdfborder={0 0 0}]{hyperref}

\usepackage[nips]{optional}

\usepackage[final]{neurips_2018}

\usepackage{array}
\usepackage{xr}
\usepackage{mathtools}
\usepackage{amsthm}
\usepackage{url}
\usepackage{comment}
\usepackage{floatrow}
\usepackage{enumitem}
\usepackage{wrapfig}
\usepackage{framed}
\usepackage{subcaption}
\usepackage{natbib}

\usepackage[capitalize]{cleveref}  
\usepackage{datetime}

\crefname{nlem}{Lemma}{Lemmas}
\crefname{nprop}{Proposition}{Propositions}
\crefname{ncor}{Corollary}{Corollaries}
\crefname{nthm}{Theorem}{Theorems}
\crefname{exa}{Example}{Examples}
\crefname{assumption}{Assumption}{Assumptions}
\crefname{equation}{}{}

\RequirePackage[usenames,dvipsnames]{color}
\usepackage{jhh-misc2}
\usepackage{mathrsfs}
\input{shortcuts}

\usepackage{autonum} 

\boldshortcuts
\suppresscomments

\graphicspath{ {figures/} }

\allowdisplaybreaks[3]

\newcommand{\dx}{\,\dee x}

\newcommand{\dz}{\,\dee z}

\newcommand{\domega}{\,\dee \omega}

\def\balign#1\ealign{\begin{align}#1\end{align}}
\def\baligns#1\ealigns{\begin{align*}#1\end{align*}}
\def\balignat#1\ealign{\begin{alignat}#1\end{alignat}}
\def\balignats#1\ealigns{\begin{alignat*}#1\end{alignat*}}
\def\bitemize#1\eitemize{\begin{itemize}#1\end{itemize}}
\def\benumerate#1\eenumerate{\begin{enumerate}#1\end{enumerate}}

\newenvironment{talign*}
 {\let\displaystyle\textstyle\csname align*\endcsname}
 {\endalign}
\newenvironment{talign}
 {\let\displaystyle\textstyle\csname align\endcsname}
 {\endalign}

\def\balignst#1\ealignst{\begin{talign*}#1\end{talign*}}
\def\balignt#1\ealignt{\begin{talign}#1\end{talign}}

\newcommand{\qtext}[1]{\quad\text{#1}\quad} 

\let\originalleft\left
\let\originalright\right
\renewcommand{\left}{\mathopen{}\mathclose\bgroup\originalleft}
\renewcommand{\right}{\aftergroup\egroup\originalright}

\def\Holder{H\"older\xspace}

\def\Matern{Mat\'ern\xspace}

\def\tinycitep*#1{{\tiny\citep*{#1}}}
\def\tinycitealt*#1{{\tiny\citealt*{#1}}}
\def\tinycite*#1{{\tiny\cite*{#1}}}
\def\smallcitep*#1{{\scriptsize\citep*{#1}}}
\def\smallcitealt*#1{{\scriptsize\citealt*{#1}}}
\def\smallcite*#1{{\scriptsize\cite*{#1}}}

\def\mbb#1{\mathbb{#1}}

\def\textsum{{\textstyle\sum}} 

\def\reals{\mathbb{R}} 

\def\naturals{\mathbb{N}} 

\def\complex{\mathbb{C}} 

\def\<{\left\langle} 
\def\>{\right\rangle}

\def\iff{\Leftrightarrow}

\def\defeq{\triangleq} 
\def\half{\frac{1}{2}}

\def\norm#1{\left\|{#1}\right\|} 
\newcommand{\onenorm}[1]{\norm{#1}_1} 
\newcommand{\twonorm}[1]{\norm{#1}_2} 
\newcommand{\infnorm}[1]{\norm{#1}_{\infty}} 
\def\staticnorm#1{\|{#1}\|} 
\newcommand{\inner}[2]{\langle{#1},{#2}\rangle} 

\def\E{\mbb{E}} 
\def\Earg#1{\E\left[{#1}\right]}
\def\Esub#1{\E_{#1}}
\def\Esubarg#1#2{\E_{#1}\left[{#2}\right]}
\def\P{\mbb{P}} 

\renewcommand{\exp}[1]{\operatorname{exp}\left(#1\right)} 

\def\independenT#1#2{\mathrel{\rlap{$#1#2$}\mkern4mu{#1#2}}}
\ifdefined\nonewproofenvironments\else
\ifdefined\ispres\else

\newenvironment{proof-sketch}{\noindent\textbf{Proof Sketch}
  \hspace*{1em}}{\qed\bigskip\\}
\newenvironment{proof-idea}{\noindent\textbf{Proof Idea}
  \hspace*{1em}}{\qed\bigskip\\}
\newenvironment{proof-of-lemma}[1][{}]{\noindent\textbf{Proof of Lemma {#1}}
  \hspace*{1em}}{\qed\\}
\newenvironment{proof-of-theorem}[1][{}]{\noindent\textbf{Proof of Theorem {#1}}
  \hspace*{1em}}{\qed\\}
\newenvironment{proof-attempt}{\noindent\textbf{Proof Attempt}
  \hspace*{1em}}{\qed\bigskip\\}

\newenvironment{remarks}{\noindent\textbf{Remarks}
  \hspace*{1em}}{\smallskip}
\fi

\theoremstyle{definition}
\newtheorem{assumption}{Assumption}
\fi

\newcommand{\pset}[0]{\mathcal{P}} 
\newcommand{\gset}[0]{\mathcal{G}} 

\newcommand{\kdnorm}[1]{\norm{#1}_{\kset_k^d}}

\newcommand{\hset}[0]{\mathcal{H}} 
\newcommand{\kset}[0]{\mathcal{K}} 

\newcommand{\blset}{BL_{\norm{\cdot}}} 
\newcommand{\twoblset}{BL_{\twonorm{\cdot}}} 

\newcommand{\operator}[1]{\mathcal{T}{#1}} 
\newcommand{\opsub}[1]{\mathcal{T}_{#1}} 
\newcommand{\opsubarg}[3]{(\opsub{#1}{#2})({#3})} 

\newcommand{\langevin}[1]{\mathcal{T}_P{#1}} 
\newcommand{\langarg}[2]{(\langevin{#1})({#2})} 

\newcommand{\twobl}{d_{\twoblset}} 

\newcommand{\qvar}[0]{x} 
\newcommand{\pvar}[0]{z}
\newcommand{\QVAR}[0]{\MakeUppercase{\qvar}} 
\newcommand{\PVAR}[0]{\MakeUppercase{\pvar}} 

\usepackage[colorinlistoftodos,prependcaption,textsize=tiny]{todonotes}

\title{Random Feature Stein Discrepancies}

\author{
Jonathan H.~Huggins\thanks{Contributed equally} \\
Department of Biostatistics, Harvard \\
\texttt{jhuggins@mit.edu}
\And 
Lester Mackey\footnotemark[1] \\
Microsoft Research New England  \\
\texttt{lmackey@microsoft.com}
}

\begin{document}

\maketitle

\begin{abstract}
Computable Stein discrepancies have been deployed for a variety of applications,
ranging from sampler selection in posterior inference to approximate Bayesian inference to goodness-of-fit testing. 
Existing convergence-determining Stein discrepancies admit strong theoretical guarantees but
suffer from a computational cost that grows quadratically in the sample size. %
While linear-time Stein discrepancies have been proposed for goodness-of-fit testing, 
they exhibit avoidable degradations in testing power---even when power is explicitly optimized.
To address these shortcomings, we introduce \emph{feature Stein discrepancies} ({\FSD}s), a new family of quality measures
that can be cheaply approximated using importance sampling.
We show how to construct {\FSD}s that provably determine the convergence of a sample to its target
and develop high-accuracy approximations---\emph{random {\FSD}s} ({\RFSD}s)---which are 
computable in near-linear time.
In our experiments with sampler selection for approximate posterior inference and goodness-of-fit testing,
{\RFSD}s perform as well or better than quadratic-time KSDs while being orders of magnitude faster to compute.
\end{abstract}

\section{Introduction}

Motivated by the intractable integration problems arising from Bayesian inference and maximum likelihood estimation~\citep{Geyer:1991}, 
\citet{Gorham:2015} introduced the notion of a Stein discrepancy as a quality measure that can potentially be computed even when direct 
integration under the distribution of interest is unavailable.  
Two classes of computable Stein discrepancies---the graph Stein discrepancy \citep{Gorham:2015,Gorham:2016b} 
and the kernel Stein discrepancy (KSD)~\citep{Oates:2017,Liu:2016b,Chwialkowski:2016,Gorham:2017}---have since been developed 
to assess and tune Markov chain Monte Carlo samplers, test goodness-of-fit, train generative adversarial networks and variational autoencoders, and 
more~\citep{Gorham:2015,Gorham:2016b,Liu:2016a,Liu:2016b,Chwialkowski:2016,Gorham:2017,Wang:2016,Liu:2017b,Jitkrittum:2017}.
However, in practice, the cost of these Stein discrepancies grows quadratically in the size of the sample being evaluated, limiting scalability.
\citet{Jitkrittum:2017} introduced a special form of KSD termed the finite-set Stein discrepancy (FSSD) to test goodness-of-fit in linear time.  However, even after an optimization-based preprocessing step to improve power, the proposed FSSD experiences a unnecessary degradation of power relative to quadratic-time tests in higher dimensions.

To address the distinct shortcomings of existing linear- and quadratic-time Stein discrepancies, 
we introduce a new class of Stein discrepancies we call \emph{feature Stein discrepancies} (\FSDs).
We show how to construct \FSDs that provably determine the convergence of a sample to its target,
thus making them attractive for goodness-of-fit testing, measuring sample quality, and other applications. 
We then introduce a fast importance sampling-based approximation we call \emph{random \FSDs} (\RFSDs). 
We provide conditions under which, with an appropriate choice of proposal distribution,
an \RFSD is close in relative error to the \FSD with high probability. 
Using an \RFSD, we show how, for any $\gamma > 0$, we can compute $O_P(N^{-1/2})$-precision estimates 
of an \FSD in $O(N^{1+\gamma})$ (near-linear) time when the \FSD precision is $\Omega(N^{-1/2})$.
Additionally, to enable applications to goodness-of-fit testing, we (1) show how to construct \RFSDs that can distinguish 
between arbitrary distributions and (2) describe the asymptotic null distribution when sample points are generated \iid\ from an unknown distribution.
In our experiments with biased Markov chain Monte Carlo (MCMC) hyperparameter selection and fast goodness-of-fit testing, we obtain high-quality 
results---which are comparable to or better than those produced by quadratic-time KSDs---using only ten features
and requiring orders-of-magnitude less computation.
\paragraph{Notation}
For measures $\mu_1, \mu_2$ on $\reals^D$ and functions ${f:\reals^D \to\complex}$, ${k: \reals^D \times \reals^D \to \complex}$, 
we let $\mu_1(f) \defined \int f(x)\mu_{1}(\dee x)$,
$(\mu_1 k)(x') \defined \int k(x, x') \mu_1(\dee x)$, and
$(\mu_{1} \times \mu_{2})(k) \defined \int \int k(x_{1}, x_{2}) \mu_{1}(\dee x_{1}) \mu_{2}(\dee x_{2})$.
We denote the generalized Fourier transform of $f$ by $\FT f$ or $\FTop(f)$ and its inverse 
by  $\invFTop(f)$.  %
For $r \ge 1$, let $\Lp{r} \defined \{ f : \staticLpnorm{r}{f} \defined (\int |f(x)|^{r}\dx)^{1/r} <\infty\}$
and $C^{n}$ denote the space of $n$-times continuously differentiable functions.
We let $\convD$ and $\convP$ denote convergence in distribution and in probability, respectively.
We let $\overline{a}$ denote the complex conjugate of $a$.
For $D \in \naturals$, define $[D] \defined \{1, \dots, D\}$. 
The symbol $\gtrsim$ indicates greater than up to a universal constant.

\section{Feature Stein discrepancies} \label{sec:methods}
When exact integration under a target distribution $P$ is infeasible, one often appeals to a 
discrete measure $\approxdist[N] = \frac{1}{N}\sum_{n=1}^N \delta_{x_n}$ to approximate expectations, 
where the sample points $x_1,\dots, x_N\in \reals^D$ are generated from a Markov chain or quadrature rule.
The aim in sample quality measurement %
is to quantify how well $\approxdist[N]$ approximates the target in a manner that
(a) recognizes when a sample sequence is converging to the target, (b) highlights when a sample sequence is not converging to the target, and (c) is computationally efficient.
It is natural to frame this comparison in terms of an integral probability metric (IPM) \citep{Muller97}, $d_\hset(\approxdist,P) \defined \sup_{h\in\hset} |\approxdist(h)-P(h)|$, measuring the maximum discrepancy between target and sample expectations over a class of test functions.
However, when generic integration under $P$ is intractable, %
standard IPMs like the $1$-Wasserstein distance and Dudley metric may not be efficiently computable.

To address this need%
, \citet{Gorham:2015} introduced the Stein discrepancy framework for generating IPM-type quality measures with no explicit integration under $P$.  
For any approximating probability measure $\mu$, each Stein discrepancy takes the form 
\[
d_{\operator{\gset}}(\mu,P) 
	= \sup_{g\in\gset} |\mu(\operator{g})| \qtext{where} \forall g \in \gset, P(\operator{g}) = 0.
\]
Here, $\operator{}$ is an operator that generates mean-zero functions under $P$, and $\gset{}$ is the \emph{Stein set} of functions on which $\operator{}$ operates.
For concreteness, we will assume that $P$ has $C^1$ density $p$ with support $\reals^d$ and restrict our attention to the popular \emph{Langevin Stein operator} \citep{Gorham:2015,Oates:2017}
defined by
$\operator{g} \defined \sum_{d=1}^D \opsub{d}{g_d}$ for $\opsubarg{d}{g_d}{x} \defined {p(x)^{-1}}{\partial_{x_d} (p(x)g_d(x))}$
and $g:\reals^D \to \reals^D$.
To date, two classes of computable Stein discrepancies with strong convergence-determining guarantees have been identified.
The graph Stein discrepancies \citep{Gorham:2015,Gorham:2016b} impose smoothness constraints on the functions $g$ and are computed by solving a linear program, while the kernel Stein discrepancies \citep{Oates:2017,Liu:2016b,Chwialkowski:2016,Gorham:2017} define $\gset$ as the unit ball of a reproducing kernel Hilbert space and are computed in closed-form.  
Both classes, however, suffer from a computational cost that grows quadratically in the number of sample points. 
Our aim is to develop alternative discrepancy measures that retain the 
theoretical and practical benefits of existing
Stein discrepancies at a greatly reduced computational cost.

Our strategy is to identify a family of convergence-determining discrepancy measures that can be accurately and inexpensively approximated with random sampling.
To this end, we define a new domain for the Stein operator centered around a \emph{feature function} $\feat : \reals^D \times \reals^D \to \complex$ 
which,  for some $r \in [1,\infty)$ and all $x, z\in\reals^D$, satisfies $\feat(x,\cdot) \in \Lp{r}$  and $\feat(\cdot, z) \in C^1$:
\balignt
\gset_{\feat,r} 
	\defined \left\{ g : \reals^{D} \to \reals \given  g_d(x) = \int \feat(x,z) \overline{f_d(z)} \dz
		\qtext{  with  } 
		\sum_{d=1}^D  \norm{f_d}_{\Lp{s}}^2 \leq 1
		\text{ for }
		s = {\frac{r}{r-1}}
		\right\}.
\ealignt
When combined with the Langevin Stein operator $\operator{}$, this \emph{feature Stein set}
gives rise to a \emph{feature Stein discrepancy} (\FSD) with an appealing explicit form $(\textsum_{d=1}^D \staticnorm{\mu(\opsub{d}{\feat})}_{\Lp{r}}^2)^{1/2}$:
\balignt
\FSD_{\feat,r}^2(\mu,\targetdist) 
	&\defined \sup_{g\in\gset_{\feat,r}} |\mu(\operator{g})|^2
	= \sup_{g\in\gset_{\feat,r}} \left|\sum_{d=1}^D \mu(\opsub{d}{g_d})\right|^2 \\ 
	&= \sup_{f : v_d = \norm{f_d}_{\Lp{s}}, \twonorm{v} \leq 1} \left|\sum_{d=1}^D \int \mu(\opsub{d}{\Phi})(z) \overline{f_d(z)} \dz\right|^2 \\
	&= \sup_{v : \twonorm{v} \leq 1} \left|\sum_{d=1}^D \norm{\mu(\opsub{d}{\Phi})}_{\Lp{r}} v_d\right|^2
	= \textsum_{d=1}^D \staticnorm{\mu(\opsub{d}{\feat})}_{\Lp{r}}^2. \label{eqn:fsd}
\ealignt
In \cref{sec:choosing-feat}, we will show how to select the feature function $\feat$ and order $r$ so that $\FSD_{\feat,r}$ provably determines convergence, in line with our desiderata (a) and (b).

To achieve efficient computation, we will approximate the \FSD in expression \cref{eqn:fsd} using an importance sample of size $M$ drawn from a proposal distribution with (Lebesgue) density $\isdist$.
We call the resulting stochastic discrepancy measure a \emph{random \FSD} ({\RFSD}):
\balignt
\RFSD_{\feat,r,\isdist,M}^2(\mu,\targetdist)
	&\defined \sum_{d=1}^D \left(\frac{1}{M}\sum_{m=1}^M{\isdist(Z_m)^{-1}}{|\mu(\opsub{d}{\feat})(Z_m)|^r}\right)^{2/r} 
	\text{ for } Z_1, \dots, Z_M \distiid \isdist.
\ealignt
Importantly, when $\mu$ is the sample approximation $\approxdist$, the \RFSD can be computed in $O(M N)$ time by evaluating the $MND$ rescaled random features, $(\opsub{d}{\feat})(x_n, Z_m)/\isdist(Z_m)^{1/r}$; this computation is also straightforwardly parallelized.
In \cref{sec:selecting-isdist}, we will show how to choose $\isdist$ so that $\RFSD_{\feat,r,\isdist,M}$ approximates $\FSD_{\feat,r}$ with small relative error.
\paragraph{Special cases}
When $r = 2$, the \FSD is an instance of a kernel Stein discrepancy (KSD) with base reproducing kernel
$\basekernel(x,y) = \int \Phi(x,z)\overline{\Phi(y,z)} \dz$. 
This follows from the definition~\citep{Oates:2017,Liu:2016b,Chwialkowski:2016,Gorham:2017}
$
\KSD_{k}(\mu,P)^2 
	\defined \textsum_{d=1}^D (\mu \times \mu)((\opsub{d} \otimes \opsub{d})\basekernel)
	= \textsum_{d=1}^D \staticnorm{\mu(\opsub{d}{\feat})}_{\Lp{2}}^2
	= \FSD_{\Phi,2}(\mu,P)^2.
$
However, we will see in \cref{sec:theory,sec:experiments} that there are significant theoretical and practical benefits to using \FSDs with $r\neq 2$.
Namely, we will be able to approximate $\FSD_{\feat,r}$ with $r\neq 2$ more effectively with a smaller sampling budget. %
If $\feat(x,z) = e^{-i\inner{z}{x}}\FT{\Psi}(z)^{1/2}$ and $\nu\propto \FT{\Psi}$ for $\Psi \in \Lp{2}$, then $\RFSD_{\feat,2,\isdist,M}$ is the random Fourier feature (RFF) approximation~\citep{Rahimi:2007} to $\KSD_{\basekernel}$ with $k(x,y) = \Psi(x-y)$. 
\citet[Prop. 1]{Chwialkowski:2015} showed that the RFF approximation can be a undesirable choice of discrepancy measure, as there exist uncountably many pairs of distinct distributions that, with high probability, cannot be distinguished by the RFF approximation.
Following \citet{Chwialkowski:2015} and \citet{Jitkrittum:2017}, we show how to select $\feat$ and $\nu$ to avoid this property in \cref{sec:asymptotics}.
The random finite set Stein discrepancy \citep[FSSD-rand,][]{Jitkrittum:2017} with proposal $\nu$ is an $\RFSD_{\feat,2,\isdist,M}$
with $\feat(x,z) = f(x,z) \nu(z)^{1/2}$ for $f$ a real analytic and $C_0$-universal \citep[Def. 4.1]{Carmeli2010} reproducing kernel.
In \cref{sec:choosing-feat}, we will see that features $\feat$ of a different form give rise to strong convergence-determining properties.

\section{Selecting a Random Feature Stein Discrepancy}
\label{sec:theory}
In this section, we provide guidance for selecting the components of an \RFSD to achieve our theoretical and computational goals.
We first discuss the choice of the feature function $\feat$ and order $r$ and then turn our attention to the proposal distribution $\isdist$.
Finally, we detail two practical choices of \RFSD that will be used in our experiments.
To ease notation, we will present theoretical guarantees in terms of the sample measure $\approxdist$, but all results continue to hold if any approximating probability measure $\mu$ is substituted for $\approxdist$.

\subsection{Selecting a feature function $\feat$} \label{sec:choosing-feat}
A principal concern in selecting a feature function is ensuring that the \FSD detects non-convergence---that is, $\approxdist \convD \targetdist$
whenever $\FSD_{\feat, r}(\approxdist, P) \to 0$.
To ensure this, we will construct \FSDs that upper bound a reference \KSD known to detect non-convergence.
This is enabled by the following inequality proved in \cref{sec:proof-fsd-lower-bound}. 
\bnprop[\KSD-\FSD inequality]\label{prop:fsd-lower-bound}
If ${k(x, y) = \int \FTop(\feat(x,\cdot))(\omega) \overline{\FTop(\feat(y,\cdot))(\omega)}\rho(\omega)\domega}$, ${r \in [1,2]}$, and $\rho\in \Lp{t}$ for $t = {r}{/(2-r)}$,  then
\balignt
\KSD_{\kernel}^2(\approxdist[N],\targetdist) 
\leq 
\textstyle\staticnorm{\rho}_{\Lp{t}}\FSD^{2}_{\feat,r}(\approxdist[N],\targetdist).  \label{eqn:mmd-ipm-bound}
\ealignt
\enprop
Our strategy is to first pick a KSD that detects non-convergence and then choose $\feat$ and $r$ such that \cref{eqn:mmd-ipm-bound} applies. 
Unfortunately, KSDs based on many common base kernels, like the Gaussian and \Matern, fail to detect non-convergence when $D > 2$ \cite[Thm. 6]{Gorham:2017}.
A notable exception is the KSD with inverse multiquadric (IMQ) base kernel.
\bexa[IMQ kernel] \label{exa:IMQ}
The IMQ kernel is given by $\IMQ{c}{\beta}(x-y) \defined (c^2 + \twonorm{x-y}^2)^\beta$, where $c > 0$ and $\beta < 0$. 
\citet[Thm. 8]{Gorham:2017} proved that when  $\beta \in (-1,0)$, KSDs with an IMQ base kernel
determine weak convergence on $\reals^D$ whenever $P\in\pset$, the set of distantly dissipative distributions for which $\grad \log p$ is Lipschitz.\footnote{We 
say $P$ satisfies \emph{distant dissipativity}~\cite{Eberle:2015,Gorham:2016b} if $\kappa_0 \defined \lim\inf_{r\to\infty} \kappa(r) > 0$ for
$
\kappa(r) = \inf \{ -2 {\inner{\grad\log p(x)-\grad\log p(y)}{x-y}}/{\twonorm{x-y}^2} : \twonorm{x-y} = r \} .
$}
\eexa
Let $\approxmean \defined \EE_{X \dist \approxdist}[X]$ denote the mean of $\approxdist$.
We would like to consider a broader class of base kernels, the form of which we summarize in the following assumption:
\begin{assumption} \label{asm:basekernel-form}
The base kernel has the form $\basekernel(x, y) = \multfeat_{N}(x)\statkernel(x - y)\multfeat_{N}(y)$ 
for $\statkernel \in C^{2}$, $\multfeat \in C^1$, and $\multfeat_N(x) \defined \multfeat(x - \approxmean)$,
where $\multfeat >0$ and $\grad \log \multfeat$ is bounded and Lipschitz. 
\end{assumption}

The IMQ kernel falls within the class defined  by \cref{asm:basekernel-form} (let $\multfeat = 1$ and $\statkernel = \IMQ{c}{\beta}$).
On the other hand, our next result, proved in \cref{sec:tilted-weak-convergence-proof}, shows that \emph{tilted base kernels} 
with $\multfeat$ increasing sufficiently quickly also control convergence. 

\begin{nthm}[Tilted KSDs detect non-convergence]\label{thm:tilted-weak-convergence}
Suppose that $\targetdist \in\pset$, \cref{asm:basekernel-form} holds, $1/\multfeat \in \Lp{2}$, and 
$H(u) \defined \sup_{\omega \in\reals^D} e^{-\twonorm{\omega}^2/(2u^2)}/\hat{\statkernel}(\omega)$ is finite for all $u > 0$.
Then for any sequence of probability measures $(\mu_N)_{N=1}^\infty$, if $\KSD_\basekernel(\mu_N, \targetdist) \to 0$
then $\mu_N \convD \targetdist$.
\end{nthm}

\bexa[Tilted hyperbolic secant kernel] \label{exa:tilted-sech}
The hyperbolic secant (sech) function is ${\sech(u) \defined 2/(e^{u} + e^{-u})}$.
For $x \in \reals^D$ and $a > 0$, define the \emph{sech kernel}
${\sechK{a}(x) \defined \prod_{d=1}^D \sech\left(\sqrt{\frac{\pi}{2}}a x_d\right)}$.
Since $\FTsechK{a}(\omega) = \sechK{1/a}(\omega)/a^D$, 
 $\KSD_\basekernel$ from \cref{thm:tilted-weak-convergence} detects non-convergence when ${\statkernel = \sechK{a}}$ and $\multfeat^{-1} \in L^2$.
Valid tilting functions include %
${\multfeat(x) = \prod_{d=1}^{D}e^{c\sqrt{1 + x_{d}^{2}}}}$ for any $c > 0$ 
and 
 $\multfeat(x) = (c^{2} + \twonorm{x}^{2})^{b}$ for any $b > D/4$ (to ensure $A^{-1} \in \Lp{2}$).
\eexa

With our appropriate reference KSDs in hand, we will now design upper bounding \FSDs.
To accomplish this we will have $\feat$ mimic the form of the base kernels in \cref{asm:basekernel-form}:
\begin{assumption} \label{asm:feature-form}
\cref{asm:basekernel-form} holds and $\feat(x, z) = \multfeat_{N}(x)\statfeat(x - z)$, where
$\statfeat \in C^{1}$ is positive, and there exist a norm $\norm{\cdot}$ and constants $s, C > 0$ such that 
\[
|\partial_{x_d} \log \statfeat(x)| \le C(1 + \norm{x}^{s}),
~~ 
\lim_{\norm{x} \to \infty}(1 + \norm{x}^{s})\statfeat(x) = 0, 
~~\text{and}~~
\statfeat(x - z) \le C \statfeat(z)/\statfeat(x).
\]
In addition, there exist a constant $\underline{c} \in (0, 1]$ and continuous, non-increasing function $\statfeatscalar$ such 
that $\underline{c}\,\statfeatscalar(\norm{x}) \le \statfeat(x) \le \statfeatscalar(\norm{x})$. 
\end{assumption}
\cref{asm:feature-form} requires a minimal amount of 
regularity from $\statfeat$, essentially that $\statfeat$ be sufficiently smooth and behave as if it is
a function only of the norm of its argument.
A conceptually straightforward choice would be to set $\statfeat = \invFTop(\FT{\statkernel}^{1/2})$---that is,
to be the \emph{square root kernel} of $\statkernel$.
We would then have that $\statkernel(x - y) = \int \statfeat(x - z)\statfeat(y - z) \dz$, so
in particular $\FSD_{\feat,2} = \KSD_{\basekernel}$. 
Since the exact square-root kernel of a base kernel can be difficult to compute in practice, 
we require only that $F$ be a suitable approximation to the square root kernel of $\statkernel$:

 \begin{assumption} \label{asm:f-is-smooth}
\cref{asm:feature-form} holds, and there exists a \emph{smoothness parameter} $\overline{\lambda} \in (1/2, 1]$ such that 
if $\lambda \in (1/2, \overline{\lambda})$, then $\FT{\statfeat}/\FT{\statkernel}^{\lambda/2} \in \Lp{2}$.
\end{assumption}
Requiring that $\FT{\statfeat}/\FT{\statkernel}^{\lambda/2} \in \Lp{2}$ is equivalent to requiring that $\statfeat$ belongs to the
reproducing kernel Hilbert space $\kset_{\lambda}$ induced by the kernel $\invFTop(\FT{\statkernel}^{\lambda})$. 
The smoothness of the functions in $\kset_{\lambda}$ increases as $\lambda$ increases.
Hence $\overline{\lambda}$ quantifies the smoothness of $\statfeat$ relative to $\statkernel$. 

Finally, we would like an assurance that the \FSD detects convergence---that is, $\FSD_{\feat, r}(\approxdist, P) \to 0$ whenever $\approxdist$
converges to $\targetdist$ in a suitable metric.
The following result, proved in  \cref{sec:ksd-upper-bound}, provides such a guarantee for both the \FSD and the \RFSD.
\bnprop \label{prop:KSD-upper-bound-for-FSD-and-RFSD}
Suppose \cref{asm:feature-form} holds with $F\in \Lp{r}$, 
$1/A$ bounded, $x \mapsto x/A(x)$ Lipschitz,  and $\Esub{P}[{A}(Z) \twonorm{Z}^2] <\infty$.
If the \emph{tilted Wasserstein distance}
\[
\textstyle
\mathcal{W}_{A_N}(\approxdist, \targetdist) \defined \sup_{h\in\hset} |\approxdist(A_Nh)-P(A_Nh)|
\quad (\hset \defined \{ h : \twonorm{\grad h(x)}\leq 1, \forall x\in\reals^D \})
\]
converges to zero, 
then $\FSD_{\feat,r}(\approxdist,\targetdist) \to 0$ and $\RFSD_{\feat,r,\isdist_{N},M_{N}}(\approxdist,\targetdist) \convP 0$ for any choices of $r\in[1,2]$, $\nu_N$, and $M_N \geq 1$.
\enprop
\bnrmk
When $A$ is constant, $\mathcal{W}_{A_N}$ is the familiar $1$-Wasserstein distance.
\enrmk

\subsection{Selecting an importance sampling distribution $\isdist$}
\label{sec:selecting-isdist}

Our next goal is to select an $\RFSD$ proposal distribution $\isdist$ for which the
\RFSD is close to its reference \FSD even when the
importance sample size $M$ is small.
Our strategy is to choose $\isdist$ so that the second moment of each \RFSD feature, 
$\wj[d](Z, \approxdist) \defined |(\approxdist\opsub{d}{\feat})(Z)|^{r}/\isdist(Z)$, is bounded by a power of its mean:
\bnumdefn[$(C,\gamma)$ second moments] \label{defn:second-moment-property}
Fix a target distribution $\targetdist$.
For $Z \dist \isdist$, $d \in [D]$, and $N \ge 1$, let $Y_{N,d} \defined \wj[d](Z, \approxdist)$. %
If for some $C > 0$ and $\gamma \in [0,2]$ we have
$
\EE[Y_{N,d}^2] \le C \EE[Y_{N,d}]^{2-\gamma}
$
for all $d \in [D]$ and $N \ge 1$,
then we say \emph{$(\feat, r, \isdist)$ yields $(C, \gamma)$ second moments for $\targetdist$ and $\approxdist$}. 
\enumdefn
The next proposition, proved in \cref{sec:estimated-ksd-lb-guarantee-proof}, demonstrates the value of this second moment property.
\bnprop \label{prop:estimated-ksd-lb-guarantee}
Suppose $(\feat,r,\isdist)$ yields $(C,\gamma)$ second moments for $P$ and $\approxdist$.
If $M \ge 2C\EE[Y_{N,d}]^{-\gamma}\log(D/\delta)/\eps^2$ for all $d \in [D]$, then,
with probability at least $1-\delta$, 
\[
\RFSD_{\feat,r,\isdist,M}(\approxdist, \targetdist)
	&\ge (1-\eps)^{1/r}\FSD_{\feat,r}(\approxdist, \targetdist).
\]
Under the further assumptions of \cref{prop:fsd-lower-bound}, if the reference $\KSD_{k}(\approxdistN, \targetdist) \gtrsim N^{-1/2}$,\footnote{Note that  $\KSD_{k}(\approxdistN,\targetdist) = \Omega_P(N^{-1/2})$ whenever the sample points $x_1, \dots, x_N$ are drawn i.i.d.\ from 
a distribution $\sourcedist$, since the scaled V-statistic $N \KSD_{k}^2(\approxdistN, \targetdist)$ diverges when $\nu \neq \targetdist$ 
and converges in distribution to a non-zero limit when $\nu = \targetdist$  \citep[Thm. 32]{Sejdinovic:2013}.
Moreover, working in a hypothesis testing framework of shrinking alternatives, \citet[Thm. 13]{Gretton:2012} 
showed that $\KSD_{k}(\approxdistN, \targetdist) = \Theta(N^{-1/2})$ was the smallest local departure distinguishable by an asymptotic KSD test.
}
then a sample size $M \gtrsim N^{\gamma r/2}C{\norm{\rho}_{\Lp{t}}^{\gamma r/2}\log(D/\delta)}{/\eps^2}$ suffices to have, with probability at least $1-\delta$,  \[
\staticnorm{\rho}_{\Lp{t}}^{1/2}\RFSD_{\feat,r,\isdist,M}(\approxdist, \targetdist)
	&\ge (1-\eps)^{1/r}\KSD_{k}(\approxdistN, \targetdist).
\]
\enprop
Notably, a smaller $r$ leads to substantial gains in the sample complexity $M = \Omega(N^{\gamma\;\!r/2})$.
For example, if $r = 1$, it suffices to choose $M = \Omega(N^{1/2})$ whenever the weight function $w_d$ is bounded (so that $\gamma = 1$);
in contrast, existing analyses of random Fourier features~\citep{Rahimi:2007,zhao2015fastmmd,sutherland2015error,Sriperumbudur:2015,Honorio:2017} require $M = \Omega(N)$ to achieve the same error rates.
We will ultimately show how to select $\isdist$ so that $\gamma$ is arbitrarily close to 0.
First, we provide simple conditions and a choice for $\isdist$ which guarantee $(C, 1)$ second moments.

\bnprop  \label{prop:c-1-second-moment}
Assume that $\targetdist \in \pset$, \cref{asm:basekernel-form,asm:feature-form} 
hold with $s=0$, and there exists a constant $\mcC' > 0$ such that for all $N \ge 1$,
$\approxdist([1 + \norm{\cdot}] \multfeat_{N}) \le \mcC'$.
If $\isdist(z) \propto \approxdist([1 + \norm{\cdot}] \feat(\cdot, z))$,
then for any $r \ge 1$, $(\feat,r,\isdist)$ yields $(C,1)$ second moments for $\targetdist$ and $\approxdist$. 
\enprop

\cref{prop:c-1-second-moment}, which is proved in \cref{sec:c-1-second-moment-proof}, 
is based on showing that the weight function $\wj[d](z, \approxdist)$ is uniformly bounded.
In order to obtain $(C, \gamma)$ moments for $\gamma < 1$, we will choose $\isdist$ such 
that $\wj[d](z, \approxdist)$ decays sufficiently quickly as $\norm{z} \to \infty$.
We achieve this by choosing an overdispersed $\isdist$---that is, we choose $\isdist$ with heavy tails compared to $\statfeat$. 
We also require two integrability conditions involving the Fourier transforms of $\statkernel$ and $\statfeat$.
\begin{assumption} \label{asm:FT-Psi-decay-plus-Lt}
\cref{asm:feature-form,asm:basekernel-form} hold, $\omega_{1}^{2}\FT{\statkernel}^{1/2}(\omega) \in \Lp{1}$, and
for $t = r/(2 - r)$, $\FT{\statkernel}/\FT{\statfeat}^{2} \in \Lp{t}$. 
\end{assumption}
The $\Lp{1}$ condition is an easily satisfied technical condition while the $\Lp{t}$ condition 
ensures that the  KSD-\FSD inequality \cref{eqn:mmd-ipm-bound} applies to our chosen \FSD.  

\bnthm \label{thm:RFSD-c-alpha-second-moment}
Assume that $\targetdist \in \pset$, \cref{asm:feature-form,asm:basekernel-form,asm:f-is-smooth,asm:FT-Psi-decay-plus-Lt} hold, and 
there exists $\mcC > 0$ such that,
\[
\approxdist\left([1 + \norm{\cdot} + \norm{\cdot - \approxmean}^{s}] \multfeat_{N}/\statfeat(\cdot - \approxmean)\right) 
\le \mcC \qtext{for all} N \ge 1. \label{eq:strong-Q-moment-bound}
\]
Then there is a constant $b \in [0, 1)$ such that the following holds.
For any $\xi \in (0, 1 - b)$, $c > 0$, and  $\alpha > 2(1 - \overline{\lambda})$, 
if $\isdist(z) \ge c\,\statkernel(z - \approxmean)^{\xi r}$, then
there exists a constant $C_{\alpha} > 0$ such that
$(\feat,r,\isdist)$ yields $(C_{\alpha}, \gamma_{\alpha})$ second moments 
for $\targetdist$ and $\approxdist$, where $\gamma_{\alpha} \defined \alpha + (2 - \alpha)\xi/(2-b-\xi)$. 
\enthm

\cref{thm:RFSD-c-alpha-second-moment} suggests a strategy for improving the importance 
sample growth rate $\gamma$ of an \RFSD: increase the smoothness $\overline{\lambda}$ of $\statfeat$ and 
decrease
the over-dispersion parameter $\xi$ of $\nu$.

\subsection{Example \RFSDs}
\label{sec:examples}

In our experiments, we will consider two \RFSDs that determine convergence by \cref{prop:fsd-lower-bound,prop:KSD-upper-bound-for-FSD-and-RFSD}
and that yield $(C, \gamma)$ second moments for any $\gamma \in (0,1]$ 
using \cref{thm:RFSD-c-alpha-second-moment}.

\bexa[$\Lp{2}$ tilted hyperbolic secant \RFSD] \label{exa:tilted-sech-RFSD}
Mimicking the construction of the hyperbolic secant kernel in \cref{exa:tilted-sech} and following
the intuition that $\statfeat$ should behave like the square root of $\statkernel$, we choose 
$\statfeat = \sechK{2a}$. 
As shown in \cref{sec:tilted-sech-RFSD-proof}, if we choose $r = 2$ 
and $\isdist(z) \propto \sechK{4a\xi}(z - \approxmean)$ we can verify all 
the assumptions necessary for \cref{thm:RFSD-c-alpha-second-moment} to hold.
Moreover, the theorem holds for any $b > 0$ and hence any $\xi \in (0, 1)$ may be chosen. 
Note that $\isdist$ can be sampled from efficiently using the inverse CDF method. 
\eexa

\bexa[$\Lp{r}$ IMQ \RFSD] \label{exa:IMQ-RFSD}
We can also parallel the construction of the reference IMQ kernel $\basekernel(x,y) = \IMQ{c}{\beta}(x-y)$
from \cref{exa:IMQ}, where $c > 0$ and $\beta \in [-D/2,0)$. 
(Recall we have $\multfeat = 1$ in \cref{asm:basekernel-form}.) 
In order to construct a corresponding \RFSD we must choose the constant $\overline{\lambda} \in (1/2,1)$
that will appear in \cref{asm:f-is-smooth} and $\underline{\xi} \in (0,1/2)$,
the minimum $\xi$ we will be able to choose when constructing $\isdist$. 
We show in \cref{sec:IMQ-RFSD-proof} that if we choose $\statfeat = \IMQ{c'}{\beta'}$,
then \cref{asm:feature-form,asm:basekernel-form,asm:f-is-smooth,asm:FT-Psi-decay-plus-Lt} hold when
$c' = \overline{\lambda} c/2$, $\beta' \in [-D/(2\underline{\xi}), -\beta/(2\underline{\xi})-D/(2\underline{\xi}))$,
$r=-D/(2\beta'\underline{\xi})$, $\xi \in (\underline{\xi}, 1)$, and $\isdist(z) \propto \IMQ{c'}{\beta'}(z - \approxmean)^{\xi r}$.
A particularly simple setting is given by $\beta' = -D/(2\underline{\xi})$, which yields $r = 1$.
Note that $\isdist$ can be sampled from efficiently since it is a multivariate $t$-distribution. 
\eexa

In the future it would be interesting to construct other {\RFSD}s.
We can recommend the following fairly simple default procedure for choosing an \RFSD based on a reference KSD admitting the form in \cref{asm:basekernel-form}.  
(1) Choose any $\gamma > 0$, and set $\alpha = \gamma/3, \bar{\lambda} = 1 - \alpha/2$, and $\xi = 4\alpha/(2 + \alpha)$. 
These are the settings we will use in our experiments. 
It may be possible to initially skip this step and reason about general choices of $\gamma$, $\xi$, and $\bar{\lambda}$. 
(2) Pick any $\statfeat$ that satisfies $\FT{\statfeat}/\FT{\statkernel}^{\lambda/2} \in \Lp{2}$ for some $\lambda \in (1/2, \bar{\lambda})$ 
(that is, \cref{asm:f-is-smooth} holds) while also satisfying $\FT{\statkernel}/\FT{\statfeat}^{2} \in \Lp{t}$ for some $t \in [1, \infty]$. 
The selection of $t$ induces a choice of $r$ via \cref{asm:FT-Psi-decay-plus-Lt}. 
A simple choice for $\statfeat$ is $\FTop^{-1}{\hat{\statkernel}^{\lambda}}$. 
(3) Check if \cref{asm:feature-form} holds (it usually does if $\statfeat$ decays no faster than a Gaussian); if it does not, a slightly different choice of $\statfeat$ should be made. 
(4) Choose $\isdist(z) \propto \statkernel(z - \approxmean)^{\xi r}$.

\begin{figure*}[tb]
\begin{center}
\begin{subfigure}[b]{0.32\textwidth} 
    \includegraphics[trim={0 0 0 0},clip,width=.9\textwidth]{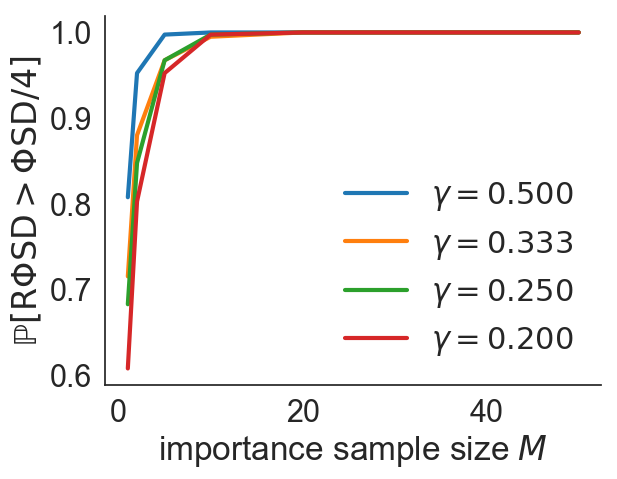} 
    \caption{Efficiency of L1 IMQ}
    \label{fig:L1-IMQ-efficiency}
\end{subfigure}  
\begin{subfigure}[b]{0.32\textwidth} 
    \includegraphics[trim={0 0 0 0},clip,width=.9\textwidth]{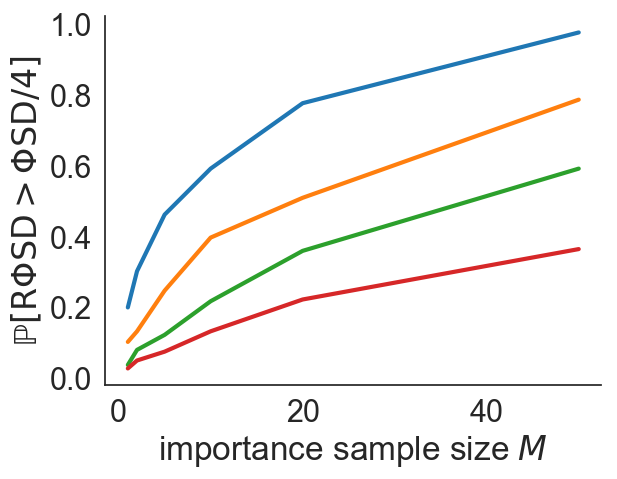} 
    \caption{Efficiency of L2 SechExp}
    \label{fig:L2-sech-efficiency}
\end{subfigure} 
\begin{subfigure}[b]{0.34\textwidth} 
    \includegraphics[trim={0 0 0 0},clip,width=.9\textwidth]{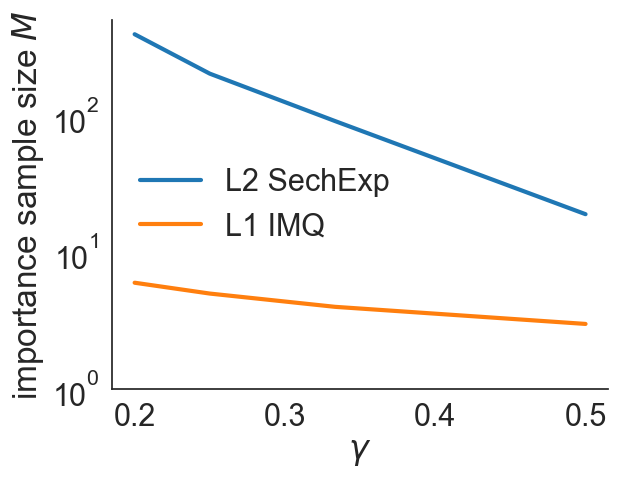} 
    \caption{$M$ necessary for $\frac{\mathrm{std}(\RFSD)}{\FSD} < \half$}
    \label{fig:sufficient-sample-sizes}
\end{subfigure} 
\end{center}
\caption{Efficiency of {\RFSD}s. The L1 IMQ \RFSD displays exceptional efficiency.}%
\label{fig:efficiency-results}
\opt{nips}{\vspace{-1em}}
\end{figure*}

\section{Goodness-of-fit testing with {\RFSDs}}\label{sec:asymptotics}
We now detail additional properties of \RFSDs relevant to testing goodness of fit.
In goodness-of-fit testing, the sample points $(X_n)_{n=1}^N$ underlying $\approxdist$ are assumed to be drawn \iid\ from a distribution $\sourcedist$,
and we wish to use the test statistic $F_{r,N} \defined \RFSD_{\feat,r,\isdist,M}^2(\approxdist,\targetdist)$
to determine whether the null hypothesis $H_{0}: \targetdist = \sourcedist$ or
alternative hypothesis $H_{1} : \targetdist \ne \sourcedist$ holds.
For this end, we will restrict our focus to real analytic $\feat$ and strictly positive analytic $\isdist$,
as by \citet[Prop.~2 and Lemmas~1-3]{Chwialkowski:2015}, with probability $1$, 
$\targetdist = \sourcedist \iff \RFSD_{\feat,r,\isdist,M}(\sourcedist, \targetdist) = 0$ when these properties hold.
Thus, analytic \RFSDs do not suffer from the shortcoming of RFFs---which are unable to distinguish
between infinitely many distributions with high probability~\citep{Chwialkowski:2015}.

It remains to estimate the distribution of the test statistic $F_{r,N}$ under the null hypothesis 
and to verify that the power of a test based on this distribution approaches 1 as $N \to \infty$.
To state our result, we assume that $M$ is fixed. 
Let  $\xi_{r,N,dm}(x) \defined (\opsub{d}\feat)(x, Z_{N,m})/(M\isdist(Z_{N,m}))^{1/r}$
for $r \in [1,2]$,
where $Z_{N,m} \distind \isdist_{N}$,
so that $\xi_{r,N}(x) \in \reals^{DM}$. 
The following result, proved in \cref{sec:asymptotics-appendix}, provides the basis for our testing guarantees.

\bnprop[Asymptotic distribution of {\RFSD}] \label{thm:RFSD-asymptotic-distribution}
Assume $\Sigma_{r,N} \defined \cov_{\targetdist}(\xi_{r,N})$ is finite for all $N$ and $\Sigma_{r} \defined \lim_{N \to \infty} \Sigma_{r,N}$ exists.
Let $\zeta \dist \distNorm(0, \Sigma_{r})$.
Then as $N \to \infty$:
(1) under $H_{0} : \targetdist = \sourcedist$, $N F_{r,N} \convD \textstyle \sum_{d=1}^{D}(\sum_{m=1}^{M}|\zeta_{dm}|^{r})^{2/r}$
and (2) under $H_{1} : \targetdist \ne \sourcedist$, $N F_{r,N} \convP \infty$.
\enprop 
\bnrmk
The condition $\Sigma_{r} \defined \lim_{N \to \infty} \Sigma_{r,N}$ holds 
if $\isdist_{N} = \isdist_{0}(\cdot - \approxmean)$ for a distribution  $\isdist_{0}$. 
\enrmk

Our second asympotic result provides a roadmap for using {\RFSD}s for hypothesis testing and is 
similar in spirit to Theorem 3 from \citet{Jitkrittum:2017}. 
In particular, it furnishes an asymptotic null distribution and establishes asymptotically full power.

\bnthm[Goodness of fit testing with {\RFSD}] \label{thm:gof-testing-with-RFSD}
Let $\hat\mu \defined N^{-1}\sum_{n=1}^{N}\xi_{r,N}(X_{n}')$ and $\hat \Sigma \defined N^{-1} \sum_{n=1}^{N}\xi_{r,N}(X_{n}')\xi_{r,N}(X_{n}')^{\top} - \hat\mu \hat\mu^{\top}$
with either $X_{n}' = X_{n}$ or $X_{n}' \distiid \targetdist$.
Suppose for the test $N F_{r,N}$, the test threshold $\tau_{\alpha}$ is set to the $(1-\alpha)$-quantile of the distribution of 
$\sum_{d=1}^{D}(\sum_{m=1}^{M} |\zeta_{dm}|^{r})^{2/r}$, where $\zeta \dist \distNorm(0, \hat\Sigma)$.
Then, under $H_{0} : \targetdist = \sourcedist$, asymptotically the false positive rate is $\alpha$.
Under $H_{1} : \targetdist \ne \sourcedist$, the test power $\Pr_{H_{1}}(N F_{r,N} > \tau_{\alpha}) \to 1$ as $N \to \infty$. 
\enthm

\begin{figure}[tb]
\begin{center}
\begin{subfigure}[b]{\textwidth} 
    \centering
    \includegraphics[trim={0 0 0 0},clip,height=90pt]{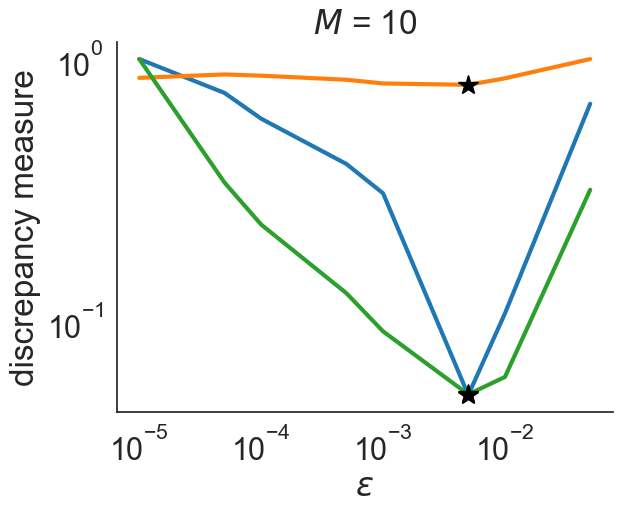}%
    \includegraphics[trim={30 0 0 0},clip,height=90pt]{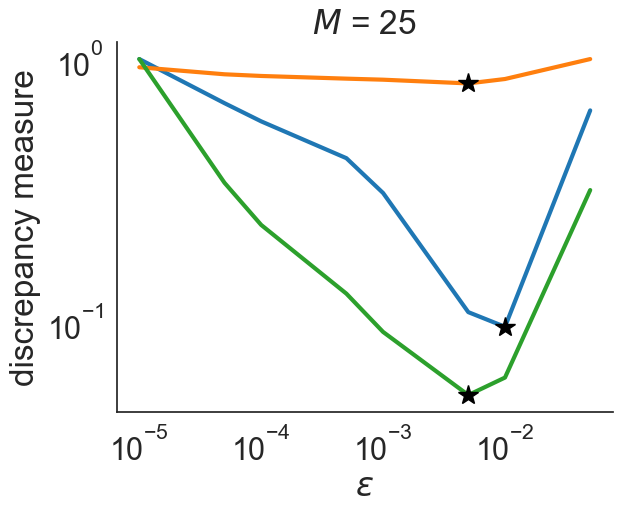}%
    \includegraphics[trim={30 0 0 0},clip,height=90pt]{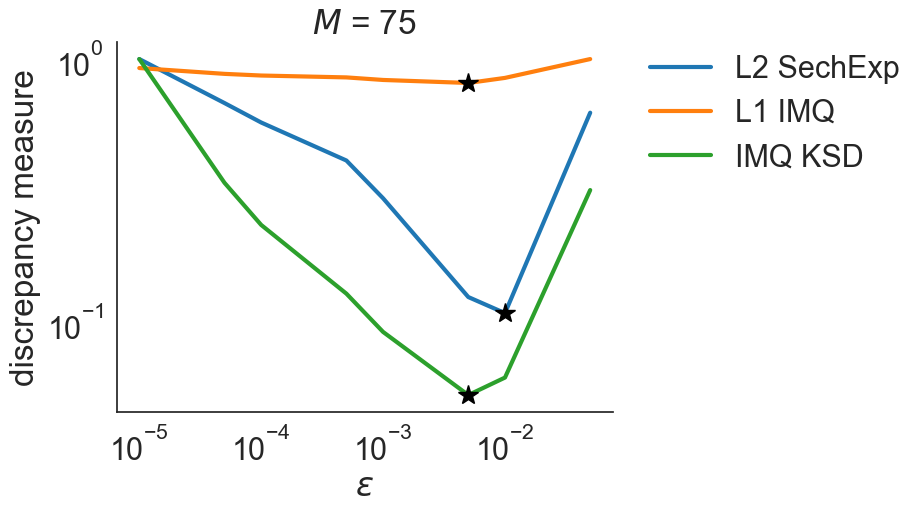}%
      \caption{Step size selection using {\RFSD}s and quadratic-time KSD baseline. %
     With $M \geq 10$, each quality measure selects a step size of $\veps =  .01$ or $.005$.}
    \label{fig:step-size-selections}
\end{subfigure}  
\begin{subfigure}[b]{\textwidth} 
   \centering
   \null\hfill\includegraphics[trim={0 0 0 0},clip,width=.23\textwidth]{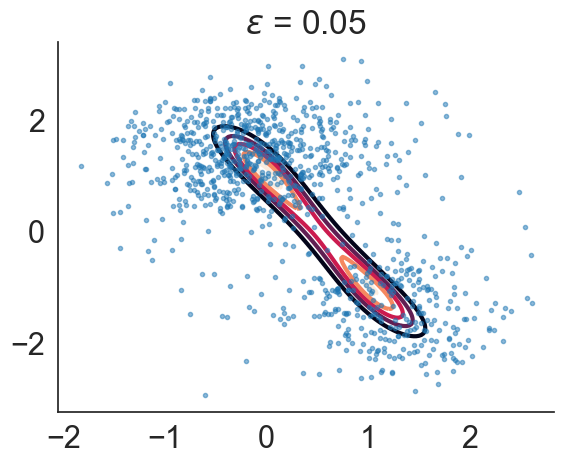}%
    \hfill\includegraphics[trim={0 0 0 0},clip,width=.23\textwidth]{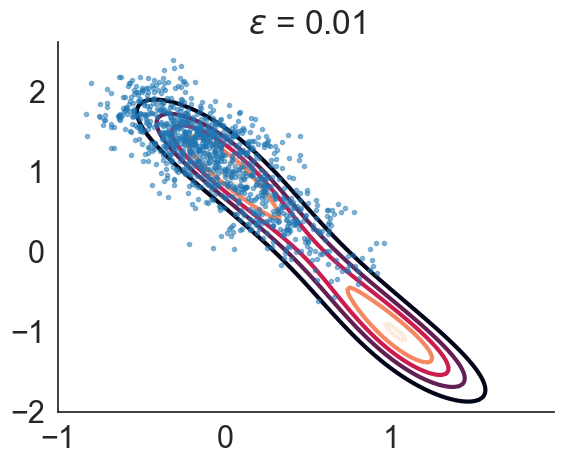}%
    \hfill\includegraphics[trim={0 0 0 0},clip,width=.23\textwidth]{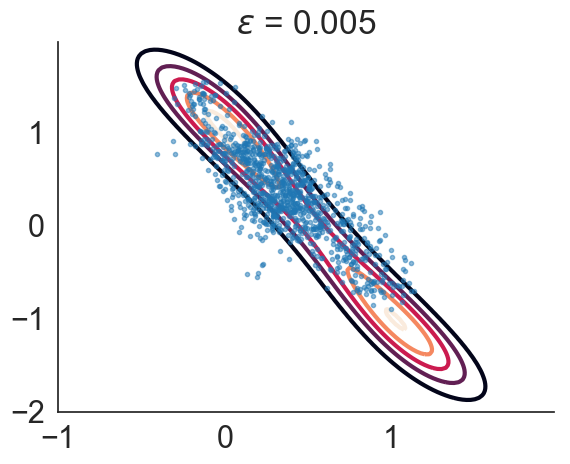}%
    \hfill\includegraphics[trim={0 0 0 0},clip,width=.23\textwidth]{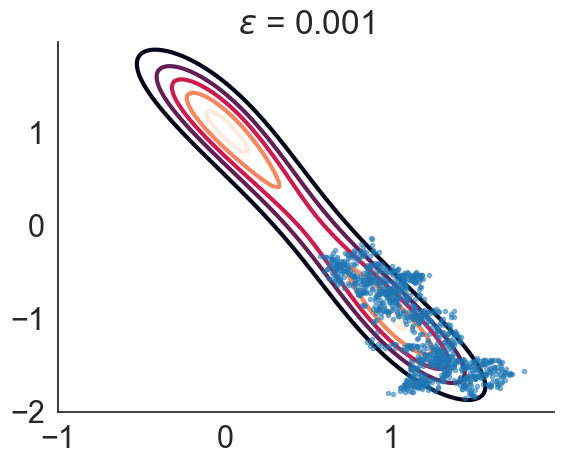}%
    \hfill\null \\ \vspace{-.75em}
     \null\hfill\includegraphics[trim={0 0 0 25},clip,width=.23\textwidth]{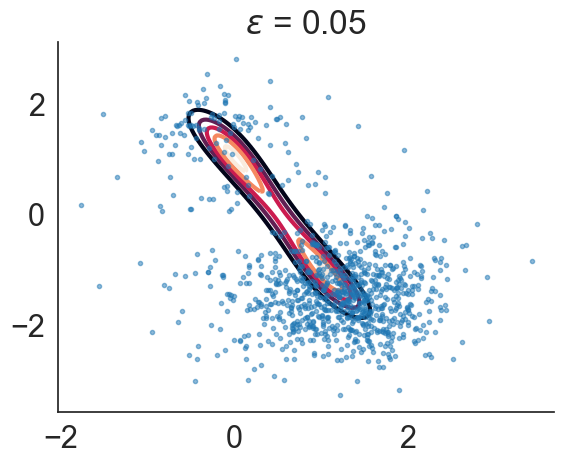}%
    \hfill\includegraphics[trim={0 0 0 25},clip,width=.23\textwidth]{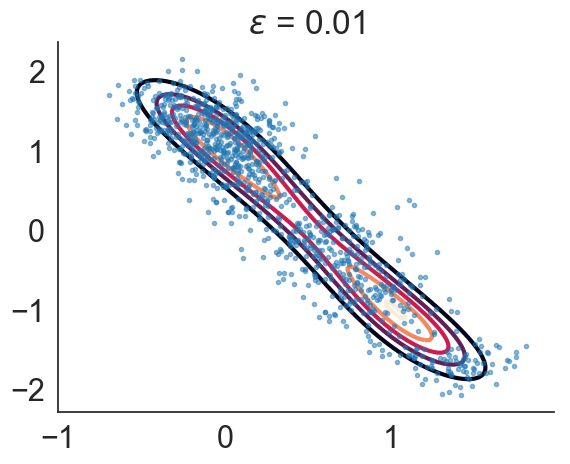}%
    \hfill\includegraphics[trim={0 0 0 25},clip,width=.23\textwidth]{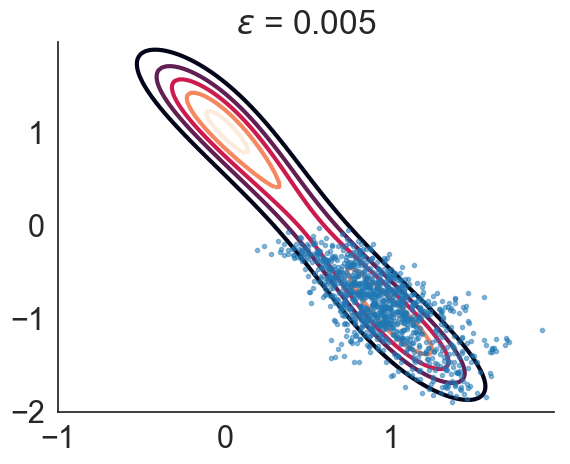}%
    \hfill\includegraphics[trim={0 0 0 25},clip,width=.23\textwidth]{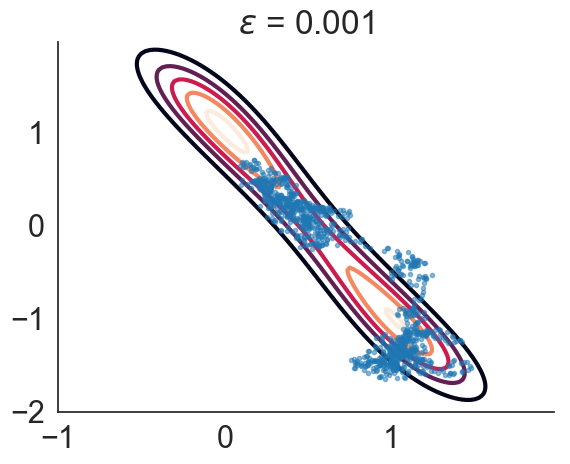}%
    \hfill\null
    \caption{SGLD sample points with equidensity contours of $\targetdensity$ overlaid. 
    The samples produced by SGLD with $\veps =  .01$ or $.005$ are noticeably better than 
    those produced using smaller or large step sizes. }
    \label{fig:sgld-samples}
\end{subfigure} 
\end{center}
\caption{Hyperparameter selection for stochastic gradient Langevin dynamics (SGLD)} %
\label{fig:sample-quality-results}
\vspace{-1em}
\end{figure}
\section{Experiments} \label{sec:experiments}

We now investigate the importance-sample and computational efficiency of our proposed {\RFSD}s 
and evaluate their benefits in MCMC hyperparameter selection and goodness-of-fit testing.\footnote{See \url{https://bitbucket.org/jhhuggins/random-feature-stein-discrepancies} for our code.}
In our experiments, we considered the {\RFSD}s described in \cref{exa:tilted-sech-RFSD,exa:IMQ-RFSD}: 
the tilted sech kernel using $r=2$ and 
$A(x) = \prod_{d=1}^{D} e^{a' \sqrt{1 + x_{d}^{2}}}$ (L2 SechExp)
and the inverse multiquadric kernel using $r=1$ (L1 IMQ). 
We selected kernel parameters as follows. 
First we chose a target $\gamma$ and then selected $\overline{\lambda}$, $\alpha$, and $\xi$ 
in accordance with the theory of \cref{sec:theory} so that
$(\feat,r,\isdist)$ yielded $(C_\gamma, \gamma)$ second moments. 
In particular, we chose $\alpha = \gamma/3$, $\overline{\lambda} = 1 - \alpha/2$, and $\xi = 4\alpha/(2 + \alpha)$.
Except for the importance sample efficiency experiments, where we varied $\gamma$ explicitly, all experiments used $\gamma  = 1/4$.
Let $\widehat{\mathrm{med}}_{u}$ denote the estimated median of the distance between data points under the $u$-norm,
where the estimate is based on using a small subsample of the full dataset. 
For L2 {SechExp}, we took $a^{-1} = \sqrt{2\pi} \widehat{\mathrm{med}}_{1}$, 
except in the sample quality experiments where we set $a^{-1} = \sqrt{2\pi}$. 
Finding hyperparameter settings for the L1 IMQ that were stable across dimension and appropriately controlled the size
for goodness-of-fit testing required some care.
However, we can offer some basic guidelines.
We recommend choosing $\underline{\xi} = \xi D / (D + df)$, which ensures $\isdist$ has $df$ degrees of freedom.
We specifically suggest using $df \in [0.5, 3]$ so that $\isdist$ is heavy-tailed no matter the dimension. 
For most experiments we took $\beta = -1/2$, $c = 4\,\widehat{\mathrm{med}}_{2}$, and $df= 0.5$. 
The exceptions were in the sample quality experiments, where we set $c = 1$, and the restricted Boltzmann machine testing experiment,
where we set $c = 10\,\widehat{\mathrm{med}}_{2}$ and $df = 2.5$.
For goodness-of-fit testing, we expect appropriate choices for $c$ and $df$ will depend on the properties of the null distribution.

\begin{wrapfigure}{R}{.42\textwidth}
\vspace{-2em}
\begin{center}
\includegraphics[trim={0 0 0 0},clip,width=\textwidth]{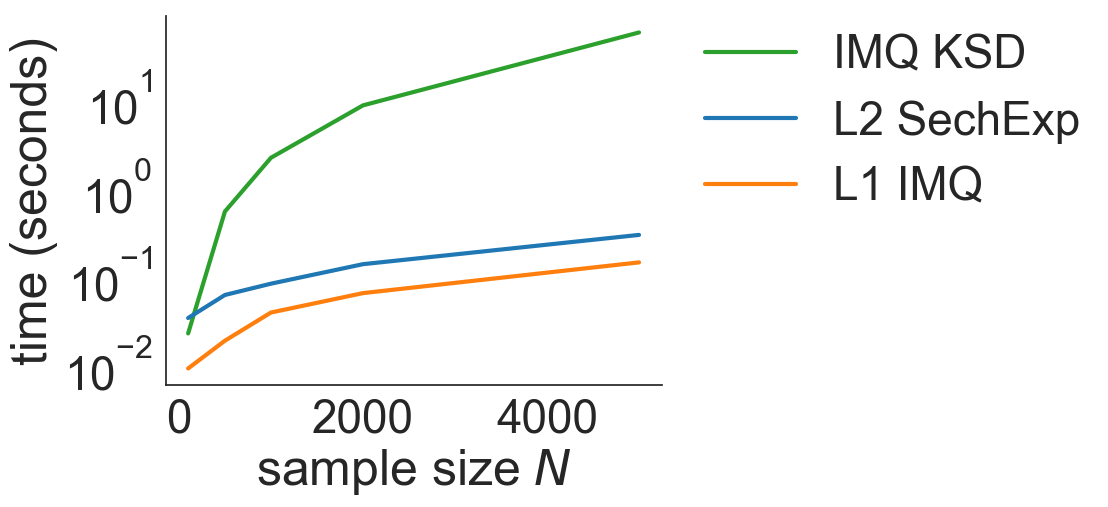}
\end{center}
\caption{Speed of IMQ KSD vs. {\RFSD}s with $M = 10$ importance sample points (dimension $D = 10$).
Even for moderate sample sizes $N$, the {\RFSD}s are orders of magnitude faster than the KSD.
}
\label{fig:speed-comparison}
\end{wrapfigure}

\paragraph{Importance sample efficiency}
\label{sec:efficiency}

To validate the importance sample efficiency theory from \cref{sec:selecting-isdist,sec:examples}, we 
calculated $\Pr[\RFSD > {\FSD}/4]$ as the importance sample size $M$ was increased.
We considered  choices of the parameters for L2 SechExp and L1 IMQ that produced
$(C_{\gamma}, \gamma)$ second moments for varying choices of $\gamma$. 
The results, shown in \cref{fig:L1-IMQ-efficiency,fig:L2-sech-efficiency}, indicate 
greater sample efficiency for L1 IMQ than L2 SechExp.
L1 IMQ is also more robust to the choice of $\gamma$.
\cref{fig:sufficient-sample-sizes}, which plots the values of $M$ necessary  
to achieve $\textrm{stdev}(\RFSD)/\FSD < 1/2$, corroborates the greater sample efficiency 
of L1 IMQ. 

\paragraph{Computational complexity}
We compared the computational complexity of the {\RFSD}s (with $M = 10$) to that of the IMQ KSD.
We generated datasets of dimension $D=10$ with the sample size $N$ ranging from $500$ to $5000$.
As seen in \cref{fig:speed-comparison}, even for moderate dataset sizes, the {\RFSD}s are computed orders of magnitude faster than the KSD.
Other {\RFSD}s like FSSD and RFF obtain similar speed-ups; however, we will see the power benefits of the L1 IMQ and L2 SechExp {\RFSD}s below.

\paragraph{Approximate MCMC hyperparameter selection}
\label{sec:hyperparameter}

We follow the stochastic gradient Langevin dynamics~\citep[SGLD, ][]{Welling:2011} hyperparameter selection setup from \citet[Section 5.3]{Gorham:2015}.
SGLD with constant step size $\veps$ is a biased MCMC algorithm that approximates the overdamped Langevin diffusion.
No Metropolis-Hastings correction is used, and an unbiased estimate of the score function from a data subsample is calculated at each iteration. 
There is a bias-variance tradeoff in the choice of step size parameter: the stationary distribution of SGLD deviates more 
from its target as $\veps$ grows, but as $\veps$ gets smaller the mixing speed of SGLD decreases. 
Hence, an appropriate choice of $\veps$ is critical for accurate posterior inference.
We target the bimodal Gaussian mixture model (GMM) posterior of \citet{Welling:2011} and compare the step size selection made
by the two {\RFSD}s to that of IMQ KSD~\citep{Gorham:2017} when $N=1000$.
\cref{fig:step-size-selections} shows that L1 IMQ and L2 SechExp agree with IMQ KSD (selecting $\veps = .005$) even with just $M=10$ importance samples.
L1 IMQ continues to select $\veps = .005$  while L2 SechExp settles on $\veps = .01$, although the value for $\veps = .005$ is only slightly larger.
\cref{fig:sgld-samples} compares the choices of $\veps = .005$ and $.01$ to smaller and larger values of $\veps$. 
The values of $M$ considered all represent substantial reductions in computation as the \RFSD replaces the $DN(N+1)/2$ KSD kernel evaluations of the form $((\opsub{d} \otimes \opsub{d})\basekernel)(x_n,x_{n'})$ with only $DNM$ feature function evaluations of the form $(\opsub{d}\Phi)(x_n,z_m)$.

\paragraph{Goodness-of-fit testing}
\label{sec:gof}

\begin{figure}[tb]
\begin{center}
\includegraphics[trim={0 0 0 0},clip,height=21pt]{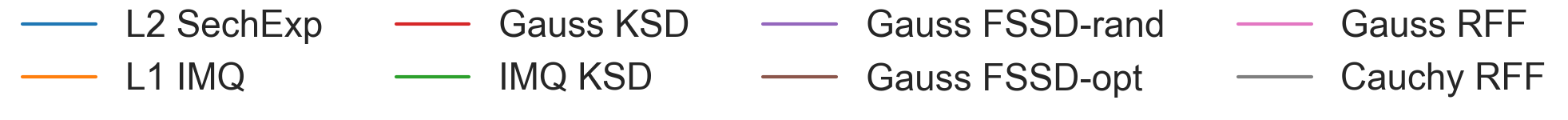} \\
\begin{subfigure}[t]{0.25\textwidth} 
    \includegraphics[trim={0 0 0 0},clip,height=70pt]{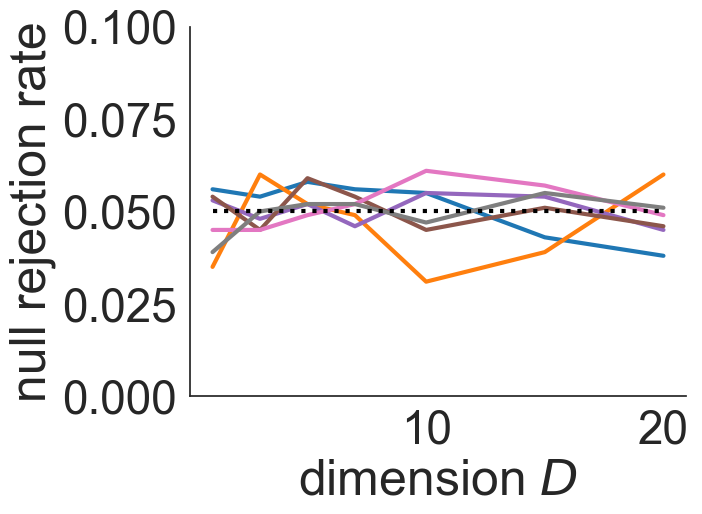}
   \caption{Gaussian null}
    \label{fig:Gaussian-vs-Gaussian-fixed-n}
\end{subfigure} 
\begin{subfigure}[t]{0.23\textwidth} 
    \includegraphics[trim={43 0 0 0},clip,height=70pt]{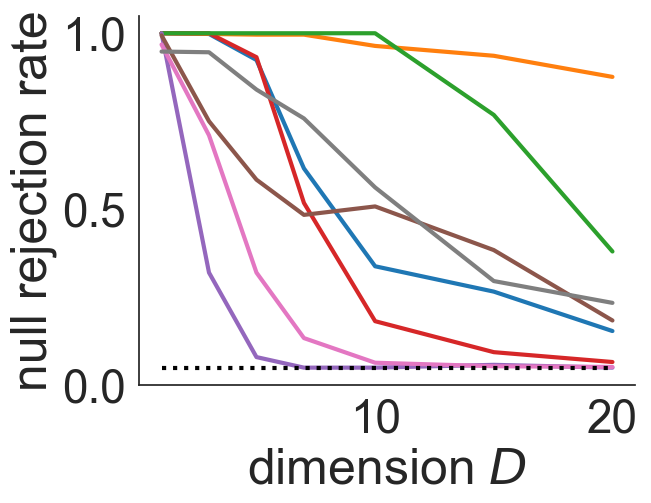}
    \caption{Gaussian vs.\ Laplace}
    \label{fig:Gaussian-vs-Laplace}
\end{subfigure}
\begin{subfigure}[t]{0.25\textwidth} 
    \includegraphics[trim={43 0 0 0},clip,height=70pt]{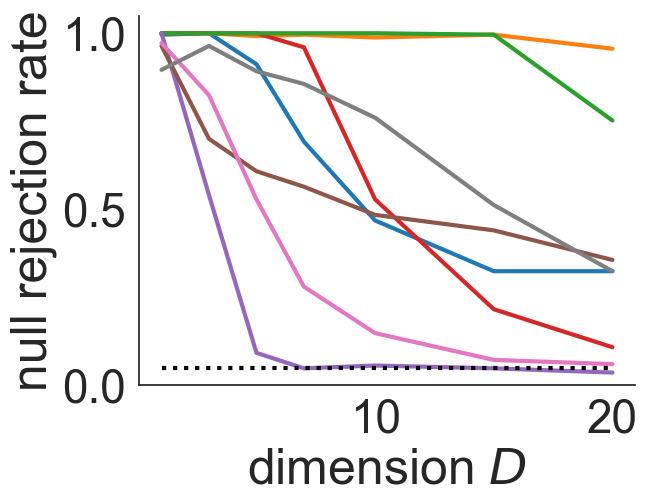} 
    \caption{Gauss vs.\ multivariate $t$}
    \label{fig:Gaussian-vs-t}
\end{subfigure}
\begin{subfigure}[t]{0.23\textwidth} 
    \includegraphics[trim={43 0 0 0},clip,height=70pt]{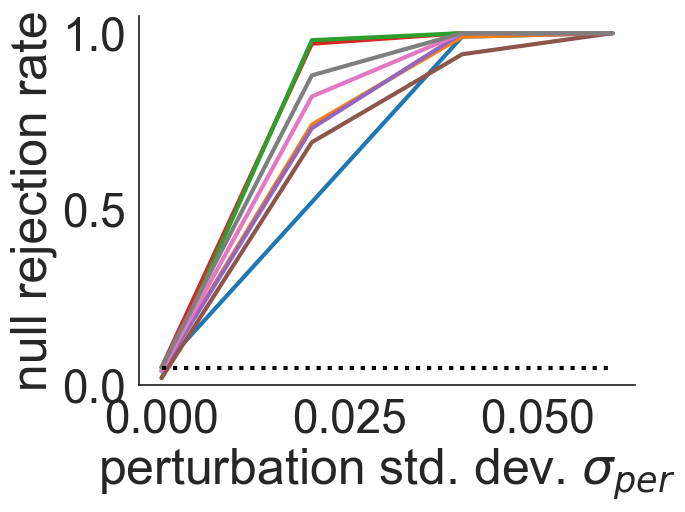} 
    \caption{RBM}
    \label{fig:RBM}
\end{subfigure}  
\end{center}
\caption{
Quadratic-time KSD and linear-time \RFSD, FSSD, and RFF goodness-of-fit tests
with $M = 10$ importance sample points %
(see \cref{sec:experiments} for more details).
All experiments used $N=1000$ except the multivariate $t$, which used $N=2000$. 
\textbf{(a)} Size of tests for Gaussian null. 
\textbf{(b, c, d)} Power of tests. Both {\RFSDs} offer competitive performance.}
\label{fig:goft-power-results}
\vspace{-.5em}
\end{figure}

Finally, we investigated the performance of {\RFSD}s for goodness-of-fit testing.
In our first two experiments we used a standard multivariate Gaussian $p(x) = \distNorm(x \given 0, I)$ as the null distribution while varying the dimension of the data.
We explored the power of {\RFSD}-based tests compared to FSSD~\citep{Jitkrittum:2017} (using the default settings in their code), RFF~\citep{Rahimi:2007} (Gaussian and Cauchy kernels with bandwidth = $\widehat{\mathrm{med}}_{2}$), 
and KSD-based tests~\citep{Chwialkowski:2016,Liu:2016b,Gorham:2017} (Gaussian kernel with bandwidth = $\widehat{\mathrm{med}}_{2}$ and IMQ kernel $\IMQ{1}{-1/2}$). 
We did not consider other linear-time KSD approximations due to relatively poor empirical performance~\citep{Jitkrittum:2017}.
There are two types of FSSD tests: FSSD-rand uses random sample locations and fixed hyperparameters while FSSD-opt uses a 
small subset of the data to optimize sample locations and hyperparameters for a power criterion. 
All linear-time tests used $M = 10$ features.
The target level was $\alpha = 0.05$.
For each dimension $D$ and {\RFSD}-based test, we chose the nominal test level by generating 200 p-values from the Gaussian asymptotic null, 
then setting the nominal level to the minimum of $\alpha$ and the 5th percentile of the generated p-values. 
All other tests had nominal level $\alpha$.
We verified the size of the FSSD, RFF, and {\RFSD}-based tests by generating 1000 p-values for each experimental setting in the Gaussian case  (see~\cref{fig:Gaussian-vs-Gaussian-fixed-n}).
Our first experiment replicated the Gaussian vs.\ Laplace experiment of \citet{Jitkrittum:2017} where, under the alternative hypothesis, $q(x) = \prod_{d=1}^{D} \distLaplace(x_{d} | 0, 1/\sqrt{2})$,  a product of Laplace distributions with variance 1 (see~\cref{fig:Gaussian-vs-Laplace}). 
Our second experiment, inspired by the Gaussian vs.\ multivariate $t$ experiment of \citet{Chwialkowski:2016}, tested the alternative in which $q(x) = \distT(x | 0, 5)$, a standard multivariate $t$-distribution with 5 degrees of freedom (see~\cref{fig:Gaussian-vs-t}). 
Our final experiment replicated the restricted Boltzmann machine (RBM) experiment of \citet{Jitkrittum:2017} in which each entry of the matrix used to
define the RBM was perturbed by independent additive Gaussian noise (see~\cref{fig:RBM}). 
The amount of noise was varied from $\sigma_{per} = 0$ (that is, the null held) up to $\sigma_{per} = 0.06$. 
The L1 IMQ test performed well across all dimensions and experiments, with power of at least 0.93 in almost all experiments.
The only exceptions were the Laplace experiment with $D = 20$ (power $\approx 0.88$) and the RBM experiment with $\sigma_{per} = 0.02$ (power $\approx 0.74$). 
The L2 SechExp test performed comparably to or better than the FSSD and RFF tests.
Despite theoretical issues, the Cauchy RFF was competitive with the other linear-time methods---except for the superior L1 IMQ. 
Given its superior power control and computational efficiency, we recommend the L1 IMQ over the L2 SechExp.

\section{Discussion and related work}

In this paper, we have introduced feature Stein discrepancies, a family of computable Stein discrepancies that can be cheaply approximated using importance sampling.
Our stochastic approximations, random feature Stein discrepancies ({\RFSD}s), combine the computational benefits of linear-time discrepancy measures with the convergence-determining properties
of quadratic-time Stein discrepancies. 
We validated the benefits of {\RFSD}s on two applications where kernel Stein discrepancies have shown excellent performance: measuring sample quality and goodness-of-fit testing. 
Empirically, the L1 IMQ {\RFSD} performed particularly well: it outperformed existing linear-time KSD approximations in high dimensions and performed as well or better than the state-of-the-art quadratic-time KSDs. 

\RFSDs could also be used as drop-in replacements for KSDs in applications to Monte Carlo variance reduction with control functionals~\citep{Oates:2017},
probabilistic inference using Stein variational gradient descent~\citep{Liu:2016a}, and kernel quadrature~\citep{Bach:2017,Briol:2017}. 
Moreover, the underlying principle used to generalize the KSD could also be used to develop fast alternatives to maximum mean discrepancies in two-sample testing applications~\citep{Gretton:2012,Chwialkowski:2015}. 
Finally, while we focused on the Langevin Stein operator, our development is compatible with any Stein operator, including diffusion Stein operators~\citep{Gorham:2016b}.

\subsubsection*{Acknowledgments}

Part of this work was done while JHH was a research intern at MSR New England.

\bibliography{library,fast_kernels}
\bibliographystyle{eabbrvnat}

\opt{nips}{
\newpage
\section*{Appendix for ``Random Feature Stein Discrepancies''}
}
\appendix
\section{Proof of {\cref{prop:fsd-lower-bound}}: KSD-\FSD inequality} \label{sec:proof-fsd-lower-bound}
We apply the generalized \Holder's inequality and the Babenko-Beckner inequality in turn to find
\[
\KSD_{\kernel}^2(\approxdist[N],\targetdist) 
	&= \textstyle\sum_{d=1}^D \int |\FTop(\approxdist(\opsub{d}{\feat}))(\omega)|^2\rho(\omega)\domega 
	\leq \textstyle\staticnorm{\rho}_{\Lp{t}}\sum_{d=1}^D \staticnorm{\FTop(\approxdist(\opsub{d}{\feat}))}_{\Lp{s}}^2 \\
	&\leq c_{r,d}^2\textstyle\staticnorm{\rho}_{\Lp{t}}\sum_{d=1}^D \staticnorm{\approxdist(\opsub{d}{\feat})}_{\Lp{r}}^2 
	= c_{r,d}^2\textstyle\staticnorm{\rho}_{\Lp{t}}\FSD^{2}_{\feat,r}(\approxdist[N],\targetdist),
\]
where $t = \frac{r}{2-r}$ and $c_{r,d} \defined (r^{1/r}/s^{1/s})^{d/2} \le 1$ for $s = r/(r-1)$.
\section{Proof of \cref{thm:tilted-weak-convergence}: Tilted KSDs detect non-convergence}\label{sec:tilted-weak-convergence-proof}
For any vector-valued function $f$, let $M_1(f) = \sup_{x,y:\twonorm{x-y}=1} \twonorm{f(x)-f(y)}$.
The result will follow from the following theorem which provides an upper bound on
the bounded Lipschitz metric $\twobl(\mu,P)$ in terms of the KSD 
and properties of $A$ and $\statkernel$.
Let $b \defined \grad \log \targetdensity$. 

\begin{nthm}[Tilted KSD lower bound]\label{thm:density-blset-upper-bound}
Suppose $\targetdist\in\pset$ and 
$\basekernel(x, y) = A(x)\statkernel(x - y)A(y)$
for $\statkernel \in C^{2}$ and $A \in C^1$ with $A >0$ and $\grad \log A$ bounded and Lipschitz. 
Then there exists a constant $\mathcal{M}_\targetdist$ such that, for all
$\eps > 0$ and all probability measures $\mu$,
\[
\twobl(\mu, \targetdist) \le \eps + C \KSD_k(\mu,\targetdist),
\]
where 
\[
C \defined (2\pi)^{-d/4} \norm{1/A}_{L^2}\mathcal{M}_\targetdist H\left (\Earg{\twonorm{G}B(G)} (1+M_1(\log A) + \mathcal{M}_\targetdist M_1(b+\grad\log A))\eps^{-1}\right )^{1/2}, 
\]
$H(t) \defined \sup_{\omega \in\reals^d} e^{-\twonorm{\omega}^2/(2t^2)}/\hat{\statkernel}(\omega)$, $G$ is a standard Gaussian vector, and $B(y) \defined \sup_{x \in \reals^d, u\in [0,1]} A(x)/A(x+uy)$.
\end{nthm}

\begin{remarks}
By bounding $H$ and optimizing over $\eps$, one can derive rates of convergence in $\twobl$.
Thm.\ 5 and Sec.\ 4.2 of \citet{Gorham:2016b} provide an explicit value for the \emph{Stein factor} $\mathcal{M}_\targetdist$.
\end{remarks}

Let $A_\mu(x) = A(x - \EE_{X \dist \mu}[X])$.
Since $\norm{1/A}_{L^2} = \norm{1/A_\mu}_{L^2}$, $M_1(\log A_\mu) \leq M_1(\log A)$, $M_1(\grad\log A_\mu) \leq M_1(\grad\log A)$, 
and $\sup_{x \in \reals^d, u\in [0,1]} A_\mu(x)/A_\mu(x+uy) = B(y)$, the exact conclusion of \cref{thm:density-blset-upper-bound}
also holds when $\basekernel(x, y) = A_\mu(x)\statkernel(x - y)A_\mu(y)$.
Moreover, since $\log A$ is Lipschitz, $B(y) \leq e^{\twonorm{y}}$ so $\Earg{\twonorm{G}B(G)}$ is finite.
Now suppose $\KSD_k(\mu_N,\targetdist)\to 0$ for a sequence of probability measures $(\mu_N)_{N\ge 1}$.
For any $\eps > 0$, $\limsup_n \twobl(\mu_N, \targetdist) \le \eps$, since $H(t)$ is finite for all $t >0$. 
Hence, $\twobl(\mu_N, \targetdist) \to 0$, and, as $\twobl$ metrizes weak convergence, $\mu_N \Rightarrow \targetdist$.

\subsection{Proof of \cref{thm:density-blset-upper-bound}: Tilted KSD lower bound}
\label{sec:density-blset-upper-bound-proof}
Our proof parallels that of \citep[Thm. 13]{Gorham:2017}.
Fix any $h\in\twoblset$.
Since $A\in C^1$ is positive, Thm.\ 5 and Sec.\ 4.2 of \citet{Gorham:2016b} imply that
there exists a $g\in C^1$ which solves the Stein equation $\langevin{(Ag)} = h - \Esubarg{P}{h(\PVAR)}$
and satisfies $M_0(Ag) \le \mathcal{M}_P$ for $\mathcal{M}_P$ a constant independent of $A, h,$ and $g$.
Since $1/A \in \Lp{2}$, we have $\norm{g}_{\Lp{2}} \leq \mathcal{M}_P \norm{1/A}_{\Lp{2}}$.

Since $\grad \log A$ is bounded, $A(x) \leq \exp{\gamma\norm{x}}$ for some $\gamma$.
Moreover, any measure in $\pset$ is sub-Gaussian, so $P$ has finite exponential moments.
Hence, since $A$ is also positive, we may define the tilted probability measure $P_A$ with density proportional to $Ap$.
The identity $\langevin{(Ag)} = A\opsub{P_A}{g}$ implies that
$$
M_0(A\grad \opsub{P_A}{g}) 
	= M_0(\grad\langevin{(Ag)} - \langevin{(Ag)} \grad \log A)
	\leq 1+M_1(\log A).
$$
Since $b$ and $\grad \log A$ are Lipschitz, we may apply the following lemma, proved in \cref{sec:stein-solution-l2-has-finite-rkhs-norm-proof} to deduce that 
there is a function $g_{\eps}\in \kset_{k_1}^d$ for $k_1(x,y)\defined \statkernel(x-y)$ such that
$|\langarg{(Ag_{\eps})}{x} - \langarg{(Ag)}{x}| = A(x)|\opsubarg{P_A}{g_{\eps}}{x} - \opsubarg{P_A}{g}{x}| \le \eps$ for all $x$ with norm
\balign\label{eqn:geps-norm-bound}
\lefteqn{\norm{g_{\eps}}_{\kset_{k_1}^d}}  \\
&\le (2\pi)^{-d/4} H\left (\Earg{\twonorm{G}B(G)} (1+M_1(\log A) + \mathcal{M}_PM_1(b+\grad\log A))\eps^{-1}\right )^{1/2} \norm{1/A}_{L^2}\mathcal{M}_P.\nonumber
\ealign
\begin{nlem}[Stein approximations with finite RKHS norm]\label{lem:stein-solution-l2-has-finite-rkhs-norm}
Consider a function $A:\reals^d\to\reals$ satisfying 
$B(y) \defined \sup_{x \in \reals^d, u\in [0,1]} A(x)/A(x+uy)$.
Suppose $g:\reals^d\to\reals^d$ is in $L^2\cap C^1$.
If $P$ has Lipschitz log density, and $k(x,y) = \statkernel(x - y)$ for $\statkernel \in C^2$ with
generalized Fourier transform $\hat \statkernel$, then for every $\eps \in (0,1]$, there is
a function $g_{\eps}:\reals^d\to\reals^d$ such that $
|\langarg{g_{\eps}}{x} - \langarg{g}{x}|\le \eps/A(x)$ for all $x\in\reals^d$ and
\[
\textstyle
	\kdnorm{g_{\eps}} \le (2\pi)^{-d/4} H\left (\Earg{\twonorm{G}B(G)} (M_0(A\grad \langevin{g}) + M_0(Ag)M_1(b))\eps^{-1}\right )^{1/2} \norm{g}_{L^2},
\]
where $H(t) \defined \sup_{\omega \in\reals^d} e^{-\twonorm{\omega}^2/(2t^2)}/\hat{\statkernel}(\omega)$ and $G$ is a standard Gaussian vector.
\end{nlem}
Since $\norm{Ag_{\eps}}_{\kset_{k}^d} = \norm{g_{\eps}}_{\kset_{k_1}^d}$, the triangle inequality and the definition of the KSD now yield
\[
|\Esubarg{\mu}{h(\QVAR)} - \Esubarg{P}{h(\PVAR)}|
  &= |\Esubarg{\mu}{\langarg{(Ag)}{\QVAR}}| \\
&  \le |\Earg{\langarg{(Ag)}{\QVAR} - \langarg{(Ag_{\eps})}{\QVAR}}| +
|\Esubarg{\mu}{\langarg{(Ag_{\eps})}{\QVAR}}| \\
  &\textstyle\le \eps
  + \norm{g_{\eps}}_{\kset_{k_1}^d}\,
  \KSD_k(\mu,P).
\]
The advertised conclusion follows by applying the bound \cref{eqn:geps-norm-bound} and taking the supremum over all
$h\in\blset$.

\subsection{Proof of \cref{lem:stein-solution-l2-has-finite-rkhs-norm}: Stein
  approximations with finite RKHS norm} \label{sec:stein-solution-l2-has-finite-rkhs-norm-proof}
Assume $M_0(A\grad \langevin{g}) + M_0(Ag) < \infty$, as otherwise the claim is vacuous.
Our proof parallels that of \citet[Lem. 12]{Gorham:2017}.
Let $Y$ denote a standard Gaussian vector with density
$\rho$.  For each $\delta \in (0,1]$, we define $\rho_{\delta}(x) =
\delta^{-d}\rho(x/\delta)$, and for any function $f$ we write
$f_{\delta}(x) \defeq \Earg{f(x + \delta Y)}$.
Under our assumptions on $h = \langevin{g}$ and $B$, the mean value theorem
and Cauchy-Schwarz imply that for each $x \in\reals^d$ there exists $u \in[0,1]$ such that
\[
|h_{\delta}(x) - h(x)| 
	&= |\Esubarg{\rho}{h(x + \delta Y) - h(x)}| 
	= |\Esubarg{\rho}{\inner{\delta Y}{\grad h(x + \delta Y u)}}| \\
	&\leq \delta M_0(A\grad \langevin{g})\, \Esubarg{\rho}{\twonorm{Y}/A(x+\delta Y u)} 
	\leq \delta M_0(A\grad \langevin{g})\, \Esubarg{\rho}{\twonorm{Y}B(Y)}/A(x).
\] 
Now, for each $x\in\reals^d$ and $\delta > 0$,
\[
h_{\delta}(x) &= \Esubarg{\rho}{\inner{b(x + \delta Y)}{g(x + \delta Y)}} + \Earg{\inner{\grad}{g(x + \delta Y)}} \qtext{and}\\
\langarg{g_{\delta}}{x} &= \Esubarg{\rho}{\inner{b(x)}{g(x + \delta Y)}} + \Earg{\inner{\grad}{g(x + \delta Y)}},
\]
so, by Cauchy-Schwarz, the Lipschitzness of $b$, and our assumptions on $g$ and $B$, 
\[
|\langarg{g_{\delta}}{x} - h_{\delta}(x)|
  &= |\Esubarg{\rho}{\inner{b(x)- b(x + \delta Y)}{g(x + \delta Y)}}| \\
  &\le \Esubarg{\rho}{\twonorm{b(x)- b(x + \delta Y)}\twonorm{g(x + \delta Y)}} \\
  &\le M_0(Ag) M_1(b)\, \delta\, \Esubarg{\rho}{\twonorm{Y}/A(x + \delta Y)}
  \le M_0(Ag) M_1(b)\, \delta\, \Esubarg{\rho}{\twonorm{Y}B(Y)}/A(x).
\]
Thus, if we fix $\eps > 0$ and define $\tilde{\eps} = \eps/(\Esubarg{\rho}{\twonorm{Y}B(Y)}(M_0(A\grad \langevin{g}) + M_0(Ag) M_1(b)))$, the triangle inequality implies
\[
|\langarg{g_{\tilde{\eps}}}{x} - \langarg{g}{x}|
  \le |\langarg{g_{\tilde{\eps}}}{x} - h_{\tilde{\eps}}(x)| + |h_{\tilde{\eps}}(x) - h(x)|
  \le \eps/A(x).
\]

To conclude, we will bound $\kdnorm{g_{\delta}}$. 
By \citet[Thm. 10.21]{Wendland:2005}, 
\[
\kdnorm{g_{\delta}}^2 
	&= (2\pi)^{-d/2}\int_{\reals^d}\frac{|\hat{g_{\delta}}(\omega)|^2}{\hat{\Phi}(\omega)}\,d\omega
	= (2\pi)^{d/2}\int_{\reals^d}\frac{|\hat{g}(\omega)|^2\hat{\rho_{\delta}}(\omega)^2}{\hat{\Phi}(\omega)}\,d\omega \\
  	&\le (2\pi)^{-d/2}\left \{\sup_{\omega\in\reals^d} \frac{e^{-\twonorm{\omega}^2\delta^2/2}}{\hat{\Phi}(\omega)}\right\} \int_{\reals^d}|\hat{g}(\omega)|^2 \, d\omega,
\]
where we have used the Convolution Theorem \citep[Thm. 5.16]{Wendland:2005}
and the identity $\hat{\rho_{\delta}}(\omega) = \hat{\rho}(\delta\omega)$.
Finally, an application of Plancherel's theorem \citep[Thm. 1.1]{herb2011plancherel} gives $\kdnorm{g_{\delta}} \le (2\pi)^{-d/4}
F(\delta^{-1})^{1/2} \norm{g}_{L^2}$. 

\section{Proof of \cref{prop:KSD-upper-bound-for-FSD-and-RFSD}} \label{sec:ksd-upper-bound}

We begin by establishing the \FSD convergence claim.  Define the target mean $\targetmean \defined \EE_{Z \dist \targetdist}[Z]$.
Since $\log A$ is Lipschitz and $A > 0$, $A_N \leq A e^{\approxmean}$ and hence $P(A_N) < \infty$ and $\Esubarg{P}{{A_N}(Z) \twonorm{Z}^2} < \infty$ for all $N$
by our integrability assumptions on $P$.

Suppose $\mathcal{W}_{A_N}(\approxdist, \targetdist) \to 0$, and, for any probability measure $\mu$ with $\mu({A_N}) < \infty$, 
define the tilted probability measure $\mu_{A_N}$ via $d\mu_{A_N}(x) = d\mu(x) {A_N}(x)$.
By the definition of $\mathcal{W}_{A_N}$, we have $|\approxdist({A_N}h) - \targetdist({A_N}h)| \to 0$ for all $h \in\hset$.
In particular, since the constant function $h(x) = 1$ is in $\hset$, we have $|\approxdist({A_N})-\targetdist({A_N})| \to 0$.
In addition, since the functions $f_N(x) = (x-\approxmean)/A_N(x)$ are uniformly Lipschitz in $N$, we have $\approxmean - \targetmean = \approxdist(f_N) - \targetdist(f_N) \to 0$
and thus $A_N \to A_P$ for $A_P(x) \defined A(x - \targetmean) > 0$.
Therefore, $\targetdist({A_N}) \to \targetdist({A_P}) > 0$, and,
as $x/y$ is a continuous function of $(x,y)$ when $y > 0$, we have 
\[
Q_{N,{A_N}}(h) - \targetdist_{{A_N}}(h) = \approxdist({A_N}h)/\approxdist({A_N}) - \targetdist({A_N}h)/\targetdist({A_N}) \to 0 
\]
and hence the $1$-Wasserstein distance $d_{\hset}(Q_{N,{A_N}}, \targetdist_{A_N}) \to 0$.

Now note that, for any $g \in \gset_{\feat/{A_N},r}$,
\[
\approxdist(\operator{{A_N}g}) 
	&= \approxdist({A_N} \opsub{\targetdist_{A_N}}{g})
	= \approxdist({A_N}) Q_{N,{A_N}}(\opsub{\targetdist_{A_N}}{g}) \\
	&= ((\approxdist({A_N}) - \targetdist({A_N})) +  \targetdist({A_N}) )Q_{N,A_N}(\opsub{\targetdist_{A_N}}{g}) \\
	&\leq
	(\mathcal{W}_{A_N}(\approxdist, \targetdist) + \targetdist({A_N}) )Q_{N,A_N}(\opsub{\targetdist_{A_N}}{g}) 
\]
where $\opsub{\targetdist_{A_N}}{}$ is the Langevin operator for the tilted measure $\targetdist_{A_N}$, defined by 
\[ 
\opsubarg{\targetdist_{A_N}}{g}{x} = \sum_{d=1}^D {(p(x){A_N}(x))^{-1}}{\partial_{x_d} (p(x){A_N}(x)g_d(x))}.
\]
Taking a supremum over $g\in\gset_{\feat/{A_N},r}$, we find
\[
\FSD_{\feat,r}(\approxdist,\targetdist) 
\leq (\mathcal{W}_{A_N}(\approxdist, \targetdist) + \targetdist({A_N}) )\FSD_{\feat/{A_N},r}(Q_{N,{A_N}},\targetdist_{A_N}) .
\]
Furthermore, since $\feat(x,z)/{A_N}(x) = F(x-z)$, \Holder's inequality implies
\[
\sup_{x\in\reals^D} \infnorm{g(x)} &\leq \norm{F}_{\Lp{r}},\\
\sup_{x\in\reals^D,d\in[D]} \infnorm{\partial_{x_d} g(x)} &\leq \norm{\partial_{x_d} F}_{\Lp{r}}, \qtext{and} \\
\sup_{x\in\reals^D,d,d'\in[D]} \infnorm{\partial_{x_d}\partial_{x_{d'}} g(x)} &\leq \norm{\partial_{x_d}\partial_{x_{d'}} F}_{\Lp{r}}
\]
for each $g\in\gset_{\feat/{A_N},r}$.
Since $\grad \log p$ and $\grad \log {A_N}$ are Lipschitz and $\Esubarg{P}{{A_N}(Z) \twonorm{Z}^2} <\infty$,
we may therefore apply \citep[Lem. 18]{Gorham:2017} to discover that $\FSD_{\feat/{A_N},r}(Q_{N,{A_N}},\targetdist_{A_N}) \to 0$ and hence $\FSD_{\feat,r}(\approxdist,\targetdist) \to 0$ whenever the $1$-Wasserstein distance $d_{\hset}(Q_{N,{A_N}}, \targetdist_{A_N}) \to 0$.

To see that $\RFSD_{\feat,r,\isdist_N,M_N}^{2}(\approxdist,\targetdist) \convP 0$ whenever $\FSD_{\feat,r}^{2}(\approxdist,\targetdist) \to 0$, first note that since $r\in[1,2]$, we may apply Jensen's inequality to obtain
\balignt
\EE[\RFSD_{\feat,r,\isdist_N,M_N}^{2}(\approxdist,\targetdist)]
	&= \EE[\sum_{d=1}^D ( \frac{1}{M}\sum_{m=1}^M{\isdist_N(Z_m)^{-1}}{|\approxdist(\opsub{d}{\feat})(Z_m)|^r})^{2/r}] \\
	&\leq \sum_{d=1}^D (\EE[\frac{1}{M}\sum_{m=1}^M{\isdist_N(Z_m)^{-1}}{|\approxdist(\opsub{d}{\feat})(Z_m)|^r}])^{2/r}\\
	&= \FSD_{\feat,r}^{2}(\approxdist,\targetdist).
\ealignt
Hence, by Markov's inequality, for any $\eps > 0$, 
\balignt
\P[\RFSD_{\feat,r,\isdist_N,M_N}^{2}(\approxdist,\targetdist) > \eps] 
	\leq {\EE[\RFSD_{\feat,r,\isdist_N,M_N}^{2}(\approxdist,\targetdist)]/}{\eps}
	\leq {\FSD_{\feat,r}^{2}(\approxdist,\targetdist)/}{\eps} \to 0,
\ealignt
yielding the result.

\section{Proof of \cref{prop:estimated-ksd-lb-guarantee} } \label{sec:estimated-ksd-lb-guarantee-proof}

To achieve the first conclusion, for each $d \in [D]$, apply \cref{cor:var-based-nonnegative-rv-concentration} with $\delta/D$ in place of $\delta$ to the random variable
\balignt
\frac{1}{M}\sum_{m=1}^M \wj[d](Z_{m}, \approxdist).
\ealignt
The first claim follows by plugging in the high probability lower bounds from \cref{cor:var-based-nonnegative-rv-concentration} 
into $\RFSD_{\feat,r,\isdist,M}^2(\approxdist,\targetdist)$ and using the union bound. 

The equality $\E[Y_d] = \FSD^r_{\feat,r}(\approxdist, \targetdist)$, the \KSD-$\FSD$ inequality of \cref{prop:fsd-lower-bound} ($\FSD^r_{\feat,r}(\approxdist, \targetdist) \geq {\KSD}^r_{k}(\approxdist, \targetdist)\norm{\rho}_{\Lp{t}}^{-r/2}$), and the assumption $\KSD_{k}(\approxdistN, \targetdist) \gtrsim N^{-1/2}$ imply that $\E[Y_d] \gtrsim N^{-r/2}\norm{\rho}_{\Lp{t}}^{-r/2}$. 
Plugging this estimate into the initial importance sample size requirement and applying the KSD-$\FSD$ inequality once more yield the second claim.

\section{Proof of \cref{prop:c-1-second-moment} } \label{sec:c-1-second-moment-proof}

It turns out that we obtain $(C,1)$ moments whenever the weight functions $\wj[d](z, \approxdist)$ are bounded.
Let $\mcQ(\feat,\isdist,C') \defined \{\approxdist \given \sup_{z,d} \wj[d](z, \approxdist) < C' \}$. 
\bnprop 
For any $C > 0$, $(\feat,r,\isdist)$ yields $(C,1)$ second moments for $\targetdist$ and $\mcQ(\feat,\isdist,C')$.
\enprop
\bprf
It follows from the definition of $\mcQ(\feat,\isdist,C)$ that
\[
\sup_{\approxdist \in \mcQ(\feat,\isdist,C)} \sup_{d, z} |(\approxdist\opsub{d}{\feat})(z)|^{r}/\isdist(z) \le C.
\]
Hence for any $\approxdist \in \mcQ(\feat,\isdist,C)$ and $d \in [D]$, $Y_{d} \le C$ \as and thus
\[
\EE[Y_{d}^{2}] \le C' \EE[Y_{d}].
\]
\eprf

Thus, to prove \cref{prop:c-1-second-moment} it suffices to have uniform bound for $\wj[d](z, \approxdist)$ for 
all $\approxdist \in \mcQ(\mcC')$. 
Let $\sigma(x) \defined 1 + \norm{x}$ and fix some $Q \in \mcQ(\mcC')$. 
Then $\isdist(z) = \approxdist(\sigma\feat(\cdot, z))/C(\approxdist)$, 
where $C(\approxdist) \defined \staticLpnorm{1}{\statfeat}\mcQ(\sigma \multfeat(\cdot - \approxmean)) \le \staticLpnorm{1}{\statfeat}\mcC'$. 
Moreover, for $c, c' > 0$ not depending on $\approxdist$,
\[
|(\approxdist\opsub{d}{\feat})(z)|^{r}
&\le  \approxdist(|\partial_{d}\log \targetdensity + \partial_{d}\log\multfeat(\cdot - \approxmean) + \partial_{d} \log \statfeat(\cdot - z)|\feat(\cdot, z))^{r}  \\
&\le c \approxdist(|1 + \norm{\cdot} + \norm{\cdot - \approxmean}^{a}|\feat(\cdot, z))^{r}  \\
&\le c' (\mcC')^{r-1} \approxdist(\sigma\feat(\cdot, z)).
\]
Thus, 
\[
\wj[d](z, \approxdist)
&= \frac{|(\approxdist\opsub{d}{\feat})(z)|^{r}}{\isdist(z)}
\le \frac{C(\mcQ) c' (\mcC')^{r-1} \approxdist(\sigma\feat(\cdot, z))}{\approxdist(\sigma\feat(\cdot, z))}
\le  c' (\mcC')^{r}\staticLpnorm{1}{\statfeat} .
\]

\section{Technical Lemmas }

\bnlem \label{lem:uniform-mmd-bound}
If $\targetdist \in \pset$, \cref{asm:feature-form,asm:basekernel-form,asm:f-is-smooth,asm:FT-Psi-decay-plus-Lt} hold, and \cref{eq:strong-Q-moment-bound} holds,
then for any $\lambda \in (1/2, \overline{\lambda})$, 
\[
|(\approxdist\opsub{d}{\feat})(z)| 
&\le C_{\lambda,\mcC} \KSD_{\kj[d]}^{2\lambda - 1}.
\]
\enlem
\bprf
Let $\varsigma_{d}(\omega) \defined (1 + \omega_{d})^{-1} \approxdist(\opsub{d}\multfeat(\cdot - \approxmean)e^{-i\omega \cdot \cdot})$.
Applying \cref{prop:dist-f-bound} with $\mcD =  \approxdist\opsub{d}\multfeat(\cdot - \approxmean)$, $h = \statfeat$, 
$\varrho(\omega) = 1 + \omega_{d}$, and $t = 1/2$ yields
\[
|(\approxdist\opsub{d}{\feat})(z)| 
&\le \norm{\statfeat}_{\FTpow{\statkernel}{\lambda}} \left((2\pi)^{-d/2} \staticLpnorm{\infty}{\varsigma_{d}}\staticLpnorm{2}{(1 + \partial_{d})\FTpow{\statkernel}{1/4}}\right)^{2-2\lambda}\norm{\approxdist\opsub{d}\feat}_{\statkernel}^{2\lambda-1} 
\]
The finiteness of $\norm{\statfeat}_{\FTpow{\statkernel}{\lambda}}$ follows from \cref{asm:f-is-smooth}.
Using $\targetdist \in \pset$, \cref{asm:basekernel-form}, and \cref{eq:strong-Q-moment-bound} we have
\[
\varsigma_{d}(\omega) 
&= (1 + \omega_{d})^{-1}\approxdist([\partial_{d}\log \targetdensity + \partial_{d}\log\multfeat(\cdot - \approxmean) - i \omega_{d}] \multfeat(\cdot - \approxmean) e^{-i \omega \cdot \cdot}) \\
&\le C \approxdist([1 + \norm{\cdot}] \multfeat(\cdot - \approxmean) \\
&\le C \mcC',
\]
so $ \staticLpnorm{\infty}{\varsigma_{d}}$ is finite.
The finiteness of $\staticLpnorm{2}{(1 + \partial_{d})\FTpow{\statkernel}{1/4}}$ follows from the Plancherel theorem and \cref{asm:FT-Psi-decay-plus-Lt}. 
The result now follows upon noting that $\norm{\approxdist\opsub{d}\feat}_{\statkernel} = \KSD_{\kj[d]}$.
\eprf

\bnlem \label{lem:h-j-bound}
If $\targetdist \in \pset$, \cref{asm:feature-form,asm:basekernel-form} hold, and \cref{eq:strong-Q-moment-bound} holds,
then for some $b \in [0,1), C_{b} > 0$,
\[
|\approxdist\opsub{d}\feat(z)| 
&\le C_{b} \statfeat(z - \approxmean)^{1-b}. 
\]
Moreover, $b=0$ if $s=0$. 
\enlem
\bprf
We have (with $C$ a constant changing line to line)
\[
|\approxdist\opsub{d}\feat(z)| 
&\le \approxdist |\opsub{d}\feat(\cdot, z)| \\
&= \approxdist(|\partial_{d}\log \targetdensity + \partial_{d}\log\multfeat(\cdot - \approxmean) + \partial_{d} \log \statfeat(\cdot - z)|\multfeat(\cdot - \approxmean)\statfeat(\cdot - z)) \\
&\le C \approxdist(|1 + \norm{\cdot} + \norm{\cdot - z}^{s}| \multfeat(\cdot - \approxmean)\statfeat(\cdot - \approxmean)^{-1}) \statfeat(z - \approxmean) \\
&\le C \approxdist(|1 + \norm{\cdot} + \norm{\cdot - \approxmean}^{s} + \norm{z - \approxmean}^{s}| \multfeat(\cdot - \approxmean)\statfeat(\cdot - \approxmean)^{-1}) \statfeat(z - \approxmean) \\
&\le C \mcC (1 + \norm{z - \approxmean}^{s}) \statfeat(z - \approxmean).  
\]
By assumption $(1 + \norm{z}^{s}) \statfeat(z) \to 0$ as $\norm{z} \to \infty$, so
for some $C_{b} > 0$ and $b \in [0, 1)$, $(1 + \norm{z - \approxmean}^{s}) \le C_{b} \statfeat(z)^{-b}$. 
\eprf

\section{Proof of \cref{thm:RFSD-c-alpha-second-moment}: $(C, \gamma)$ second moment bounds for {\RFSD} }  \label{sec:second-moment-bounds}

Take $\approxdist \in \mcQ(\mcC)$ fixed and let $\wj[d](z) \defined \wj[d](z, \approxdist)$. 
For a set $S$ let $\isdist_S(S') \defined \int_{S \cap S'} \isdist(\dee z)$.
Let $K \defined \{ x \in \reals^{D} \given \norm{x  - \approxmean} \le R \}$.
Recall that $Z \dist \isdist$ and $Y_{d} = \wj[d](Z)$. 
We have 
\[
\EE[Y_{d}^{2}]
&=\EE[\wj[d](Z)^{2}] 
= \EE[\wj[d](Z)^{2} \ind(Z \in K)] + \EE[\wj[d](Z)^{2} \ind(Z \notin K)] \\
\begin{split}
&\le \staticLpnormarg{1}{\nu}{\wj[d]} \staticLpnormarg{\infty}{\nu}{\wj[d] \ind(\cdot \in K)} + \staticLpnormarg{1}{\nu}{\ind(\cdot \notin K)} \staticLpnormarg{\infty}{\nu}{\wj[d]^{2}\ind(\cdot \notin K)} 
\end{split} \\
&= \staticLpnorm{r}{\approxdist\opsub{d}{\feat}}^{r} \sup_{z \in K}\wj[d](z) + \nu(K^\complement) \sup_{z \in K^\complement}\wj[d](z)^{2} \\
&= \EE[Y_{d}] \sup_{z \in K}\wj[d](z) + \nu(K^\complement) \sup_{z \in K^\complement}\wj[d](z)^{2}
\]
Without loss of generality we can take $\isdist(z) = \statkernel(z - \approxmean)^{\xi r}/\staticLpnorm{1}{\statkernel^{\xi r}}$, 
since a different choice of $\isdist$ only affects constant factors. 
Applying \cref{lem:uniform-mmd-bound,asm:FT-Psi-decay-plus-Lt,eqn:mmd-ipm-bound}, we have
\[
\sup_{z \in K}\wj[d](z)
&\le C_{\lambda,\mcC}^{r}\KSD_{\kj[d]}^{r(2\lambda - 1)} \sup_{z \in K} \isdist(z)^{-1} \\
&\le  C_{\lambda,\mcC}^{r}\staticLpnorm{1}{\statkernel^{\xi r}}\sup_{z \in K} \statfeat(z - \approxmean)^{-\xi r} \KSD_{\kj[d]}^{r(2\lambda - 1)} \\
&\le  C_{\lambda,\mcC}^{r}\underline{c}^{-\xi r}\staticLpnorm{1}{\statkernel^{\xi r}}\staticLpnorm{t}{\FT{\statkernel}/\FT{\statfeat}^{2}}\statfeatscalar(R)^{-\xi r} \staticnorm{\approxdist\opsub{d}{\feat}}_{\Lp{r}}^{r(2\lambda - 1)} \\
&=  C_{\lambda,\mcC}^{r}\staticLpnorm{1}{(\statkernel/\underline{c})^{\xi r}}\staticLpnorm{t}{\FT{\statkernel}/\FT{\statfeat}^{2}}\statfeatscalar(R)^{-\xi r} \EE[Y_{d}]^{2\lambda - 1}.
\]
Applying \cref{lem:h-j-bound} we have 
\[
\sup_{z \in K^\complement}\wj[d](z)^{2}
&\le C_{b}^{2} \sup_{z \in K^\complement}\statfeat(z - \approxmean)^{2(1 - b)r}/\isdist(z)^{2} \\
&=  C_{b}^{2}\staticLpnorm{1}{\statkernel^{\xi r}}^{2}\sup_{z \in K^\complement}\statfeat(z - \approxmean)^{2(1 - b - \xi)r} \\
&= C_{b}^{2}\staticLpnorm{1}{\statkernel^{\xi r}}^{2} \statfeatscalar(R)^{2(1 - b - \xi)r}.
\]
Thus, we have that 
\[
\EE[Y_{d}^{2}]
&\le C_{\lambda,\mcC,r,\xi} \EE[Y_{d}]^{2\lambda} \statfeatscalar(R)^{-\xi r} + C_{b,\xi r}  \statfeatscalar(R)^{2(1 - b - \xi)r}.
\]
As long as $\EE[Y_{d}]^{2\lambda} \le C_{b,\xi r}\statfeatscalar(0)^{2(1 - b - \xi/2)r}/ C_{\lambda,\mcC,r,\xi}$,
since $\statfeatscalar$ is continuous and non-increasing to zero we can choose $R$ such that 
$\statfeatscalar(R)^{2(1 - b - \xi)r} =  C_{\lambda,\mcC,r,\xi} \EE[Y_{d}]^{2\lambda}/ C_{b,\xi r}$
and the result follows for $$\EE[Y_{d}]^{2\lambda} \le C_{b,\xi r}\statfeatscalar(0)^{2(1 - b - \xi/2)r}/ C_{\lambda,\mcC,r,\xi}.$$
Otherwise, we can guarantee that $\EE[Y_{d}^{2}] \le C_{\alpha}\EE[Y_{d}]^{2-\gamma_{\alpha}}$ be choosing 
$C_{\alpha}$ sufficiently large, since by assumption $\EE[Y_{d}]$ is uniformly bounded over 
$\approxdist \in \mcQ(\mcC)$. 
\section{A uniform MMD-type bound }  \label{sec:uniform-mmd-bound}

Let $\mcD$ denote a tempered distribution and $\statkernel$ a stationary kernel. 
Also, define $\hat\mcD(\omega) \defined \mcD_{x}e^{-i \inner{\omega}{\hx}}$.

\bnprop \label{prop:dist-f-bound}
Let $h$ be a symmetric function such that for some $s \in (0, 1]$, $h \in \kset_{\FTpow{\statkernel}{s}}$ 
and ${\mcD_{x}h(\hx - \cdot) \in \kset_{\FTpow{\statkernel}{s}}}$. %
Then
\[
|\mcD_{x}h(\hx - z)|
&\le \norm{h}_{\FTpow{\statkernel}{s}}\norm{\mcD_{x}\FTpow{\statkernel}{s}(\hx - \cdot)}_{\FTpow{\statkernel}{s}}
\]
and for any $t \in (0, s)$ any function $\varrho(\omega)$, 
\[
\norm{\mcD_{x}\FTpow{\statkernel}{s}(\hx - \cdot)}_{\FTpow{\statkernel}{s}}^{1-t}
&\le \left((2\pi)^{-d/2} \Lpnorm{\infty}{\varrho^{-1}\hat\mcD}\Lpnorm{2}{\varrho\FT{\statkernel}^{t/2}}\right)^{1-s}\norm{\mcD_{x}\statkernel(\hx - \cdot)}_{\statkernel}^{s-t}.
\]
Furthermore, if for some $c > 0$ and $r \in (0, s/2)$, ${\FT{h} \le c\,\FT{\statkernel}^{r}}$, then
\[
\norm{h}_{\FTpow{\statkernel}{s}} &\le \frac{c \Lpnorm{2}{\FTpow{\statkernel}{r - s/2}}}{(2\pi)^{d/4}}.
\]
\enprop
\bprf
The first inequality follows from an application of Cauchy-Schwartz:
\[
|\mcD_{x}h(\hx - z)|
&= |\inner{h(\cdot - z)}{\mcD_{x} \FTpow{\statkernel}{s}(\hx - \cdot)}_{\FTpow{\statkernel}{s}}| \\
&\le  \norm{h(\cdot - z)}_{\FTpow{\statkernel}{s}}\norm{\mcD_{x} \FTpow{\statkernel}{s}(\hx - \cdot)}_{\FTpow{\statkernel}{s}} \\
&=  \norm{h}_{\FTpow{\statkernel}{s}}\norm{\mcD_{x} \FTpow{\statkernel}{s}(\hx - \cdot)}_{\FTpow{\statkernel}{s}}.
\]
For the first norm, we have
\[
\norm{h}_{\FTpow{\Phi}{s}}^{2}
&= (2\pi)^{-d/2}\int \frac{\FT{h}^{2}(\omega)}{\FT\Phi^{s}(\omega)}\,\dee\omega  \\
&\le c^{2}(2\pi)^{-d/2} \int\FT\Phi^{2r - s}(\omega)\,\dee\omega \\
&= c^{2}(2\pi)^{-d/2}\Lpnorm{2}{\FTpow{\statkernel}{r - s/2}}^{2}. 
\]
Note that by  the convolution theorem $\FTop(\mcD_{x} \FTpow{\statkernel}{s}(\hx - \cdot))(\omega) = \hat\mcD(\omega)\FT\statkernel^{s}(\omega)$. 
For the second norm, applying Jensen's inequality and \Holder's inequality yields
\[
\norm{\mcD_{x}\FTpow{\statkernel}{s}(\hx - \cdot)}_{\FTpow{\statkernel}{s}}^{2}
&= (2\pi)^{-d/2}\int \frac{\FT\statkernel(\omega)^{2s}|\hat\mcD(\omega)|^{2}}{\FT\statkernel^{s}(\omega)}\,\dee\omega  \\
&= (2\pi)^{-d/2}\left(\int\FT\statkernel^{t}|\hat\mcD|^{2}\right)\int \frac{\FT\statkernel(\omega)^{t}|\hat\mcD(\omega)|^{2}}{\int\FT\statkernel^{t}|\hat\mcD|^{2}} \FT\statkernel(\omega)^{s-t}\,\dee\omega \\
&\le (2\pi)^{-d/2}\left(\int\FT\statkernel^{t}|\hat\mcD|^{2}\right)\left(\int\frac{\FT\statkernel(\omega)^{t}|\hat\mcD(\omega)|^{2}}{\int\FT\statkernel^{t}|\hat\mcD|^{2}}\statkernel(\omega)^{1-t}\,\dee\omega \right)^{\frac{s-t}{1-t}} \\
&=\left((2\pi)^{-d/2}\int\FT\statkernel^{t}|\hat\mcD|^{2}\right)^{\frac{1-s}{1-t}}\norm{\mcD_{x}\statkernel(\hx - \cdot)}_{\statkernel}^{2\frac{s-t}{1-t}} \\
&\le\left( (2\pi)^{-d/2}\Lpnorm{\infty}{|\varrho^{-1}\hat\mcD|^2}\int \varrho^{2}\FT\statkernel^{t} \right)^{\frac{1-s}{1-t}}\norm{\mcD_{x}\statkernel(\hx - \cdot)}_{\statkernel}^{2\frac{s-t}{1-t}} \\
&= \left( (2\pi)^{-d/2}\Lpnorm{\infty}{\varrho^{-1}\hat\mcD}^{2} \Lpnorm{2}{\varrho\FT{\statkernel}^{t/2}}^{2}\right)^{\frac{1-s}{1-t}}\norm{\mcD_{x}\statkernel(\hx - \cdot)}_{\statkernel}^{2\frac{s-t}{1-t}}.
\]
\eprf

\section{Verifying \cref{exa:tilted-sech-RFSD}: Tilted hyperbolic secant {\RFSD} properties }
\label{sec:tilted-sech-RFSD-proof}

We verify each of the assumptions in turn. 
By construction or assumption each condition in \cref{asm:basekernel-form} holds. 
Note in particular that $\sechK{2a} \in C^{\infty}$. 
Since $e^{-a |x_{d}|} \le \sech(a x_{d}) \le 2 e^{-a |x_{d}|}$, \cref{asm:feature-form} holds 
with $\norm{\cdot} = \onenorm{\cdot}$, $\statfeatscalar(R) = 2^{d}e^{- \sqrt{\frac{\pi}{2}} a R}$,
and $\underline{c} = 2^{-d}$, and $s = 1$.
In particular,
\[
\partial_{x_{d}} \log \sechK{2a}(x)
&= \sqrt{2\pi}a\tanh(\sqrt{2\pi}a x_{d}) + \textsum_{d' \ne d}^{D} \log \sech(\sqrt{2\pi}a x_{d'}) \\
&\le (\sqrt{2\pi}a)(1 + \textsum_{d' \ne d}^{D} |x_{d'}) \\
&\le (\sqrt{2\pi}a)(1 + \onenorm{x})
\]
and using \cref{prop:sech-kernel-self-bound} we have that 
\[
\sechK{a}(x - z) 
\le e^{\sqrt{\frac{\pi}{2}}a \onenorm{x}}\sechK{a}(z) 
\le 2^{d}\sechK{a}(z) /\sechK{a}(x).
\]

\cref{asm:f-is-smooth} holds with $\overline{\lambda} = 1$ since for any $\lambda \in (0,1)$, 
it follow from \cref{prop:fractional-sech-power-bounds} that 
\[
\wideFT{\gj}/\FTstatKernj^{\lambda/2}
= \FTsechK{2a}/(\FTsechK{a})^{\lambda/2}
\le 2^{d/2}(\FTsechK{2a})^{1-\lambda}
\in \Lp{2}.
\] 
The first part of \cref{asm:FT-Psi-decay-plus-Lt} holds as well since by \cref{eq:sech-FT},
$\omega_{d}^{2}\FTsechK{a}(\omega) = a^{-D}\omega_{d}^{2}\sechK{1/a}(\omega) \in \Lp{1}$. 

Finally, to verify the second part of \cref{asm:FT-Psi-decay-plus-Lt}, we first note that since $r = 2$, $t = \infty$. 
The assumption holds since by \cref{prop:fractional-sech-power-bounds}, 
$\FTsechK{a}(\omega)/\FTsechK{2a}(\omega)^{2} \le 1$.

\section{Verifying \cref{exa:IMQ-RFSD}: IMQ {\RFSD} properties }
\label{sec:IMQ-RFSD-proof}

We verify each of the assumptions in turn. 
By construction or assumption each condition in \cref{asm:basekernel-form} holds. 
Note in particular that $\IMQ{c'}{\beta'} \in C^{\infty}$. 
\cref{asm:feature-form} holds with $\norm{\cdot} = \twonorm{\cdot}$, $\statfeatscalar(R) = ((c')^{2} + R^{2})^{\beta'}$,
 $\underline{c} = 1$, and $s = 0$. 
 In particular,
\[
|\partial_{x_{d}} \log   \IMQ{c'}{\beta'}(x)|
&\le  -\frac{2\beta'|x_{d}|}{(c')^{2} + \twonorm{x}^{2}} 
\le -2\beta'
\]
and 
\[
\frac{\IMQ{c'}{\beta'}(x - z)}{\IMQ{c'}{\beta'}(z)}
 &= \left(\frac{(c')^{2} + \twonorm{x - z}^{2}}{(c')^{2} + \twonorm{z}^{2}}\right)^{-\beta'}\\
 &\le \left(\frac{(c')^{2} + 2\twonorm{z}^{2} + 2\twonorm{x}^{2}}{(c')^{2} + \twonorm{z}^{2}}\right)^{-\beta'} \\
 &\le \left(2 + 2\twonorm{x}^{2}/(c')^{2}\right)^{-\beta'} \\
 &= 2^{-\beta} \IMQ{c'}{\beta'}(x)^{-1}.
\]

By \citet[Theorem 8.15]{Wendland:2005}, $\IMQ{c}{\beta}$ has generalized Fourier transform
\[%
\widehat{\IMQ{c}{\beta}}(\omega) = \frac{2^{1+\beta}}{\Gamma(-\beta)}
\left(\frac{\twonorm{\omega}}{c}\right )^{-\beta-D/2} K_{\beta + D/2}(c\twonorm{\omega}),
\]
where $K_v(z)$ is the modified Bessel function of the third kind.
We write $a(\ell) \asympequivconst b(\ell)$ to denote asymptotic equivalence up to a constant: $\lim_{\ell} a(\ell)/b(\ell) = c$ for some $c \in (0,\infty)$. 
Asymptotically~\citep[eqs.~10.25.3, 
10.30.2]{NIST:DLMF},%
\[
\FTIMQ{c}{\beta}(\omega) 
	&\asympequivconst  \twonorm{\omega}^{-\beta-D/2-1/2} e^{-c \twonorm{\omega}}, & \twonorm{\omega} &\to \infty \qtext{and}\\
\FTIMQ{c}{\beta}(\omega)
	&\asympequivconst \twonorm{\omega}^{-(\beta+D/2)-|\beta+D/2|}  =  \twonorm{\omega}^{-(2\beta+D)_+} & \twonorm{\omega} &\to 0 .
\]
\cref{asm:f-is-smooth} holds since for any $\lambda \in (0,\overline{\lambda})$,
\[
\FTIMQ{c'}{\beta'} / (\FTIMQ{c}{\beta})^{\lambda/2}
	&\asympequiv  \twonorm{\omega}^{-(\beta'+D/2-1/2)+(\beta+D/2-1/2)\lambda/2} e^{(-c' +c\lambda/2) \twonorm{\omega}}, & \twonorm{\omega} &\to \infty \qtext{and}\\
	&\asympequiv \twonorm{\omega}^{\lambda(2\beta+D)_+/2-(2\beta'+D)_+} = \twonorm{\omega}^{\lambda(2\beta+D)/2} & \twonorm{\omega} &\to 0, 
\]
so $\FTIMQ{c'}{\beta'} / (\FTIMQ{c}{\beta})^{\lambda/2} \in \Lp{2}$ as long as 
$c' = c\overline{\lambda}/2 > c\lambda/2$ and $\lambda(2\beta+D) > -D$.
The first condition holds by construction and  second condition is always satisfied, since $2\beta+D \geq 0 > -D$. 

The first part of \cref{asm:FT-Psi-decay-plus-Lt} holds as well since $\FTIMQ{c'}{\beta'}(\omega)$ decreases exponentially as $\twonorm{\omega} \to \infty$ 
and $\FTIMQ{c'}{\beta'}(\omega) \asympequiv 1$ as $\twonorm{\omega} \to 0$, so $\omega_{d}^{2}\FTIMQ{c'}{\beta'}(\omega)$ 
is integrable.

Finally, to verify the second part of \cref{asm:FT-Psi-decay-plus-Lt} we first note that $t = r/(2-r) = -D/(D+4\beta'\underline{\xi})$. 
Thus,
\[
\FTIMQ{c}{\beta}/ (\FTIMQ{c'}{\beta'})^{2} 
	&\asympequivconst \twonorm{\omega}^{-2(\beta+D/2-1/2)/2+2(\beta'+D/2-1/2))} e^{2(-c/2+c') \twonorm{\omega}}, & \twonorm{\omega} &\to \infty \qtext{and}\\
	&\asympequivconst \twonorm{\omega}^{2(2\beta'+D)_+-(2\beta+D)_+}
	= \twonorm{\omega}^{-(2\beta+D)}
	& \twonorm{\omega} &\to 0,
\]
so $\FTIMQ{c}{\beta}/ (\FTIMQ{c'}{\beta'})^{2}  \in \Lp{t}$ whenever $c/2>c'$ and 
\[
\frac{D}{(D+4\beta'\underline{\xi})}(2\beta+D) > -D 
\iff -\beta/(2\underline{\xi})-D/(2\underline{\xi}) > \beta'.
\]
Both these conditions hold by construction.

\section{Proofs of \cref{thm:RFSD-asymptotic-distribution,thm:gof-testing-with-RFSD}: Asymptotics of {\RFSD}}  \label{sec:asymptotics-appendix}

The proofs of \cref{thm:RFSD-asymptotic-distribution,thm:gof-testing-with-RFSD} rely on the following asymptotic result. 
\bnthm \label{thm:gof-estimator-asymptotics}
Let $\xi_{i} : \reals^{D} \times \mcZ \to \reals, i=1,\dots,I$, be a collection of functions; 
let $Z_{N,m} \distind \isdist_{N}$, where $\isdist_{N}$ is a distribution on $\mcZ$;
and let $X_{n} \distiid \mu$, where $\mu$ is absolutely continuous with respect to Lebesgue measure.
Define the random variables $\xi_{N,nim} \defined \xi_{i}(X_{n}, Z_{N,m})$ and, for $r, s \ge 1$, the random variable
\[
\textstyle
F_{r,s,N} \defined \left(\sum_{i=1}^{I}\left(\sum_{m=1}^{M} \left|N^{-1} \sum_{n=1}^{N}\xi_{N,nim}\right|^{r}\right)^{s/r}\right)^{2/s}. 
\]
Assume that for all $N \ge 1$, $i \in [I]$, and $m \in [M]$, $\xi_{N,1im}$ has a finite second moment
that that $\Sigma_{im,i'm'} \defined \lim_{N \to \infty}\cov(\xi_{N,im}, \xi_{N,i'm'}) < \infty$  exists for all $i, i \in [I]$ 
and $m, m' \in [M]$. 
Then the following statements hold. 
\benum
\item If $\varrho_{N,im} \defined (\mu \times \isdist_{N})(\xi_{i}) = 0$ for all $i \in [N]$ then
\[
\textstyle
N F_{r,s,N} \convD  \left(\sum_{i=1}^{I}\left(\sum_{m=1}^{M} \left|\zeta_{im}\right|^{r}\right)^{s/r}\right)^{2/s} \text{ as } N \to \infty, \label{eq:asympotic-null-dist}
\]
where $\zeta \dist \distNorm(0, \Sigma)$.
\item If $\varrho_{N,im} \ne 0$ for some $i$ and $m$, then
\[
\textstyle
N F_{r,s,N} \convas \infty \text{ as } N \to \infty.
\]
\eenum
\enthm
\bprf
Let $V_{N,im} = N^{-1/2} \sum_{n=1}^{N} \xi_{N,nim}$. 
By assumption $\|\Sigma\| < \infty$. 
Hence, by the multivariate CLT,
\[
\textstyle
V_{N} - N^{1/2}\varrho_{N} \convD \distNorm(0, \Sigma).
\]
Observe that $N F_{r,s,N} = (\sum_{i=1}^{I}(\sum_{m=1}^{M}|V_{N,im}|^{r})^{s/r})^{2/s}$.
Hence if $\varrho = 0$, \cref{eq:asympotic-null-dist} follows from the continuous mapping theorem.

Assume $\varrho_{N,ij} \ne 0$ for some $i$ and $j$ and all $N \ge 0$.
By the strong law of large numbers,
$
N^{-1/2}V_{N} \convas \varrho_{\infty}.
$
Together with the continuous mapping theorem conclude that
$
F_{r,s,N} \convas c
$
for some $c > 0$.
Hence $N F_{r,s,N} \convas \infty$. 
\eprf

When $r = s =2$, the {\RFSD} is a degenerate $V$-statistic, and we recover its well-known distribution \citep[Sec. 6.4, Thm. B]{Serfling:1980} as a corollary. 
A similar result was used in \citet{Jitkrittum:2017} to construct the asymptotic null for the FSSD, which is degenerate $U$-statistic. 
\bncor
Under the hypotheses of \cref{thm:gof-estimator-asymptotics}(1), 
\[
\textstyle
N F_{2,2,N} \convD  \sum_{i=1}^{I}\sum_{m=1}^{M} \lambda_{im}\omega_{im}^{2} \text{ as } N \to \infty,
\]
where $\lambda = \operatorname{eigs}(\Sigma)$ and $\omega_{ij} \distiid \distNorm(0, 1)$. 
\encor

To apply these results to {\RFSD}s, take $s = 2$ and apply \cref{thm:gof-estimator-asymptotics} with $I = D$, $\xi_{N,dm} = \xi_{r,N,dm}$. 
Under $H_{0} : \mu = \targetdist$, $\targetdist(\xi_{r,N,dm}) = 0$ for all $d \in [D]$ and $m \in [M]$, so part 1 of  \cref{thm:gof-estimator-asymptotics} holds.
On the other hand, when $\mu \ne \targetdist$, there exists some $m$ and $d$ for which $\mu(\xi_{r,dm}) \ne 0$. 
Thus, under $H_{1} : \mu \ne \targetdist$ part 2 of \cref{thm:gof-estimator-asymptotics} holds.

The proof of \cref{thm:gof-testing-with-RFSD} is essentially identical to that of \citet[Theorem 3]{Jitkrittum:2017}.

\section{Hyperbolic secant properties }

Recall that the hyperbolic secant function is given by $\sech(a) = \frac{2}{e^{a} + e^{-a}}$.
For $x \in \reals^d$, define the hyperbolic secant kernel
\[
\sechK{a}(x) \defined \sech\left(\sqrt{\frac{\pi}{2}}ax\right) \defined \prod_{i=1}^d \sech\left(\sqrt{\frac{\pi}{2}}ax_i\right).
\]
It is a standard result that 
\[
\FTsechK{a}(\omega) = a^{-D}\sechK{1/a}(\omega). \label{eq:sech-FT}
\]

We can relate $\sechK{a}(x)^\xi$ to $\sechK{a\xi}(x)$, but to do so we will need the following standard result:
\bnlem \label{lem:fractional-power-bounds}
For $a, b \ge 0$ and $\xi \in (0,1]$,
\[
\frac{a^\xi + b^\xi}{2^{1-\xi}} \le (a + b)^\xi \le a^\xi + b^\xi.
\]
\enlem
\bprf
The lower bound follows from an application of Jensen's inequality and the upper bound follows from the concavity of 
$a \mapsto a^\xi$. 
\eprf

\bnprop \label{prop:fractional-sech-power-bounds}
For $\xi \in (0,1]$, 
\[
\sechK{a}(x)^\xi &\le \sechK{a}(\xi x) = \sechK{a\xi}(x) \le 2^{d(1-\xi)} \sechK{a}(x)^\xi \\
2^{-d(1-\xi)}\FTsechK{a/\xi}(x) &\le \FTsechK{a}(x)^\xi \le \FTsechK{a/\xi}(x).
\]
Thus, $\sechK{a/\xi}$ is equivalent to $\FTpow{(\sechK{a})}{\xi}$. 
\enprop
\bprf
Apply \cref{lem:fractional-power-bounds,eq:sech-FT}.
\eprf

\bnprop \label{prop:sech-kernel-self-bound}
For all $x, y \in \reals^d$ and $a > 0$, 
\[
\sechK{a}(x - z) \le e^{\sqrt{\frac{\pi}{2}}a \onenorm{x}}\sechK{a}(z).  
\]
\enprop
\bprf
Take $d=1$ since the general case follows immediately. 
Without loss of generality assume that $x \ge 0$ and
let $a' = \sqrt{\frac{\pi}{2}}a$. 
Then
\[
\frac{\sechK{a}(x - z)}{\sechK{a}(z)}
&= \frac{e^{a'z} + e^{-a'z}}{e^{a'(x-z)} + e^{-a'(x-z)}}
=  \frac{e^{a'z} + e^{-a'z}}{e^{-a'z} + e^{2 a' x} e^{a'z}} e^{a'x}
\le e^{a'x}. 
\]
\eprf
\section{Concentration inequalities } \label{sec:concentration}

\bnthm[{\citet[Theorem 2.9]{Chung:2006}}] \label{thm:lower-bounded-rv-concentration}
Let $X_1,\dots,X_m$ be independent random variables satisfying $X_i > -A$ for all $i=1,\dots,m$.
Let $X \defined \sum_{i=1}^m X_i$ and $\overline{X^2} \defined \sum_{i=1}^m \EE[X_i^2]$.
Then for all $t > 0$, 
\[
\Pr(X \le \EE[X] - t) \le e^{-\frac{1}{2}t^2/(\overline{X^2} + At/3)}.
\] 
\enthm

Let $\hX \defined \frac{1}{m}\sum_{i=1}^m X_i$. 

\bncor \label{cor:var-based-nonnegative-rv-concentration}
Let $X_1,\dots,X_m$ be \iid\ nonnegative random variables with mean $\bar X \defined \EE[X_1]$.
Assume there exist $c > 0$ and $\gamma \in [0,2]$ such that 
$\EE[X_1^2] \le c {\bar X}^{2-\gamma}$.
If, for $\delta \in (0,1)$ and $\veps \in (0,1)$,
\[
m \ge \frac{2c\log(1/\delta)}{\veps^2}{\bar X}^{-\gamma},
\]
then with probability at least $1 - \delta$,
$
\hX \ge (1 - \veps)\bar X. 
$
\encor
\bprf
Applying \cref{thm:lower-bounded-rv-concentration} with $t = m\veps\bar X$ and $A = 0$ yields
\[
\Pr(\hX \le (1-\veps) \bar X) 
	\le e^{-\frac{1}{2}\veps^2 m  \bar{X}^{2}/ (c\EE[X_1^2])}
	\le e^{-\frac{1}{2c}\veps^2 m  \bar{X}^{\gamma}}.
\]
Upper bounding the right hand side by $\delta$ and solving for $m$ yields the result. 
\eprf

\bncor \label{cor:var-based-nonnegative-rv-concentration-2}
Let $X_1,\dots,X_m$ be \iid\ nonnegative random variables with mean $\bar X \defined \EE[X_1]$.
Assume there exists $c > 0$ and $\gamma \in [0,2]$ such that 
$\EE[X_1^2] \le c {\bar X}^{2-\gamma}$.
Let $\eps' = |X^{*} - \bar X|$ and assume $\eps' \le \eta X^{*}$ for some $\eta \in (0,1)$. 
If, for $\delta \in (0,1)$,
\[
m \ge \frac{2c\log(1/\delta)}{\veps^2}{\bar X}^{-\gamma},
\]
then with probability at least $1 - \delta$, $\hX \ge (1 - \veps)X^{*}$.
In particular, if $\eps' \le \frac{\sigma X^{*}}{\sqrt{n}}$ and $X^{*} \ge \frac{\sigma^{2}}{\eta^{2}n}$, then
with probability at least $1 - \delta$, $\hX \ge (1 - \veps)X^{*}$
as long as 
\[
m \ge \frac{2c(1-\eta)^{2}\eta^{2\gamma}}{\veps^2\sigma^{2\gamma}\log(1/\delta)}n^{\gamma}.
\]
\encor
\bprf
Apply \cref{cor:var-based-nonnegative-rv-concentration} with $\frac{\veps X^{*}}{\bar{X}}$ in place of $\veps$.
\eprf
\bexa
If we take $\gamma = 1/4$ and $\eta = \veps = 1/2$, then $X^{*} \ge \frac{4\sigma^{2}}{n}$
and $m \ge \frac{\sqrt{2}c \log(1/\delta)}{\sigma^{1/2}}n^{1/4}$ guarantees that 
$\hX \ge \frac{1}{2}X^{*}$ with probability at least $1-\delta$. 
\eexa

\end{document}